\documentclass[]{bytedance_seed}

\usepackage[toc,page,header]{appendix}

\usepackage{minitoc}
\usepackage{graphicx}
\usepackage{amsmath}
\usepackage{amssymb}
\usepackage{mathtools}
\usepackage{amsthm}
\usepackage{hyperref}
\usepackage{multirow}
\usepackage{xcolor}
\usepackage{colortbl}
\usepackage{booktabs}

\usepackage{algorithm}
\usepackage{algorithmicx}
\usepackage{algpseudocode}
\usepackage{verbatim}
\usepackage{caption}
\usepackage{subcaption}
\usepackage{fancyvrb}
\usepackage{fvextra}
\usepackage{wrapfig}
\usepackage{xurl}

\definecolor{GREEN}{HTML}{62b197}
\definecolor{RED}{HTML}{e18e6d}

\DeclareUnicodeCharacter{FF5C}{\textbar}

\title{AgentGym-RL: \\
Training LLM Agents for Long-Horizon Decision Making through Multi-Turn Reinforcement Learning}

\author[1 * \dag]{Zhiheng Xi}
\author[1 *]{ Jixuan Huang}
\author[1 *]{Chenyang Liao}
\author[1]{\\ Baodai Huang}
\author[1]{Honglin Guo}
\author[1]{Jiaqi Liu}
\author[1]{Rui Zheng}
\author[1]{Junjie Ye}
\author[1]{\\ Jiazheng Zhang}
\author[1]{Wenxiang Chen}
\author[1]{Wei He}
\author[1]{Yiwen Ding}
\author[1]{Guanyu Li}
\author[2]{\\ Zehui Chen}
\author[2]{Zhengyin Du}
\author[2]{Xuesong Yao}
\author[2]{Yufei Xu}
\author[2]{Jiecao Chen}
\author[1,3 \dag ]{\\ Tao Gui}
\author[1,3]{Zuxuan Wu}
\author[1 \dag]{Qi Zhang}
\author[1]{Xuanjing Huang}
\author[1]{Yu-Gang Jiang}
\affiliation[1]{Fudan University}
\affiliation[2]{ByteDance Seed}
\affiliation[3]{Shanghai Innovation Institute}

\abstract{
Developing autonomous LLM agents capable of making a series of intelligent decisions to solve complex, real-world tasks is a fast-evolving frontier. Like human cognitive development, agents are expected to acquire knowledge and skills through exploration and interaction with the environment. Despite advances, the community still lacks a unified, interactive reinforcement learning (RL) framework that can effectively train such agents from scratch—without relying on supervised fine-tuning (SFT)—across diverse and realistic environments. To bridge this gap, we introduce AgentGym-RL, a new framework to train LLM agents for multi-turn interactive decision-making through RL. The framework features a modular and decoupled architecture, ensuring high flexibility and extensibility. It encompasses a wide variety of real-world scenarios, and supports mainstream RL algorithms. Furthermore, we propose ScalingInter-RL, a training approach designed for exploration-exploitation balance and stable RL optimization. In early stages, it emphasizes exploitation by restricting the number of interaction, and gradually shifts towards exploration with larger horizons to encourage diverse problem-solving strategies. In this way, the agent develops more diverse behaviors and is less prone to collapse under long horizons. We perform extensive experiments to validate the stability and effectiveness of both the AgentGym-RL framework and ScalingInter-RL approach. Our agents match or surpass commercial models on 27 tasks across diverse environments. We offer key insights and will open-source complete AgentGym-RL framework—including code and datasets—to empower the research community in developing the next generation of intelligent agents.
\\
\\
}

\correspondence{\email{zhxi22@m.fudan.edu.cn}, \email{tgui@fudan.edu.cn}, \email{qz@fudan.edu.cn} }
\checkdata[Code]{\url{https://github.com/woooodyy/AgentGym-RL}}
 \checkdata[Project]{\url{https://AgentGym-RL.github.io}}

\begin{document}
\maketitle

\begin{figure*}[!h]
    \centering
    \begin{subfigure}{0.58\textwidth}
        \centering
        \includegraphics[width=\linewidth]{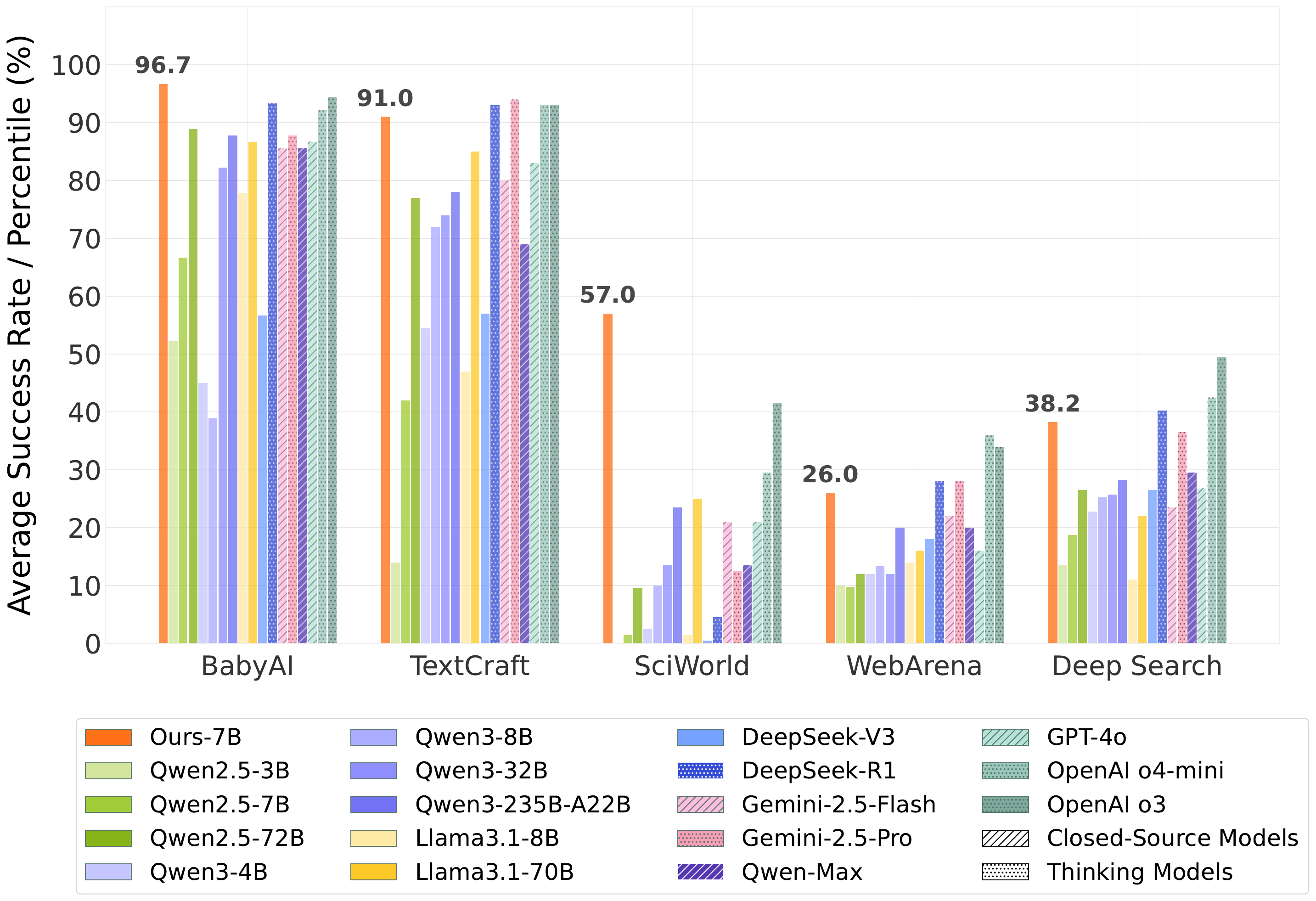}
        \caption{} 
        \label{fig:main_greedy_performance}
    \end{subfigure}%
    \hspace{0.5em}
    \begin{subfigure}{0.38\textwidth}
        \centering
        \includegraphics[width=\linewidth]{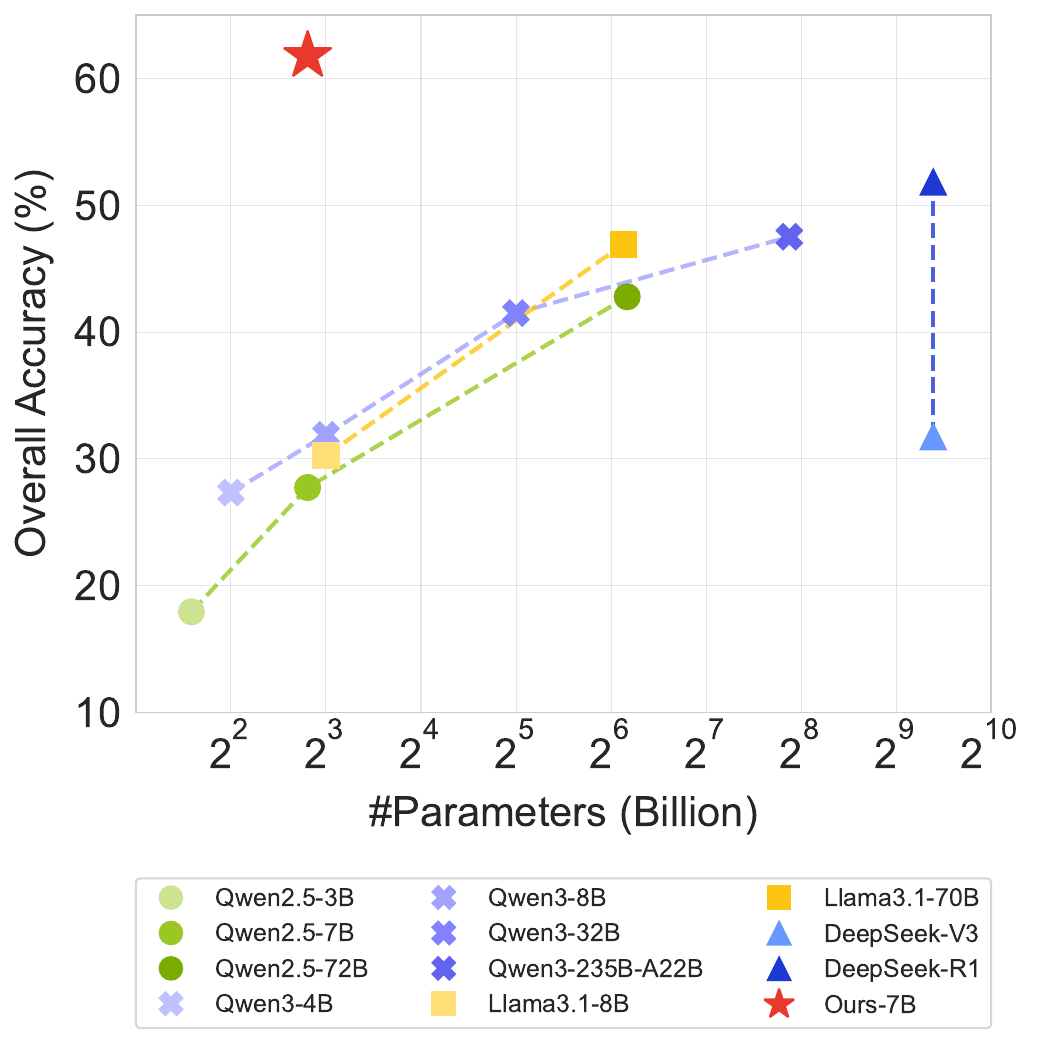}
        \caption{} 
        \label{fig:model_performance_scatter}
    \end{subfigure}
    \vspace{-10pt}
    \caption{ \textbf{Left}: Performance of proprietary models, open-source models, and our RL models across different agentic tasks. 
    \textbf{Right}: Performance w.r.t model scale. Working in concert, our framework and method substantially enhances the open-sourced 7B-scale models' capabilities to a level that rivals or even surpasses top-tier proprietary large models. }
    \label{fig:combined_performance}
\end{figure*}

\section{Introduction}

As Large Language Models (LLMs) have largely developed \citep{DBLP:journals/corr/abs-2303-08774, anthropic2024claude, DBLP:journals/corr/abs-2412-19437, team2023gemini, DBLP:journals/corr/abs-2505-09388}, their applications have extended from chatbots to autonomous agents that can handle long-horizon real-world tasks \citep{agentsurvey, moonshot2025kimik2}. Given a complex task, these agents interact with the environment, making a series of intelligent decisions to achieve the goal \citep{DBLP:conf/icml/ZhouZPLK24}. Analogous to human cognitive development, LLM agents are expected to acquire new knowledge and skills by actively exploring and interacting with the environment \citep{AgentGym, openai202x_o3o4mini}. Therefore, a natural approach is to train these agents using Reinforcement Learning (RL) \citep{RL/SuttonB2018}.

Despite the progress of RL in areas like LLM reasoning \citep{DeepSeek-R1, openai_o1, r3_2024, ReFT, kimi_k1.5, Skywork}, most existing studies are restricted to single-turn tasks, where models are not required to engage in multi-turn interaction with complex environments \citep{RAGEN}. While some recent efforts have attempted to extend RL to train LLM agents with multi-turn capabilities \citep{DBLP:conf/icml/ZhouZPLK24, RLOO, RAGEN, WebRL, Search-R1, cao2025skyrl}, these works are limited in task complexity and environment diversity. Furthermore, they struggle with optimization stability and efficiency, resulting in suboptimal performance. Critically, the community currently lacks a unified, end-to-end, interactive multi-turn RL framework that is proven to be effective across a wide range of real-world scenarios and environments for training LLM agents without SFT as a preliminary step \citep{DeepSeek-R1}.

\begin{figure*}[t]
    \centering
    \includegraphics[width=0.99\linewidth]{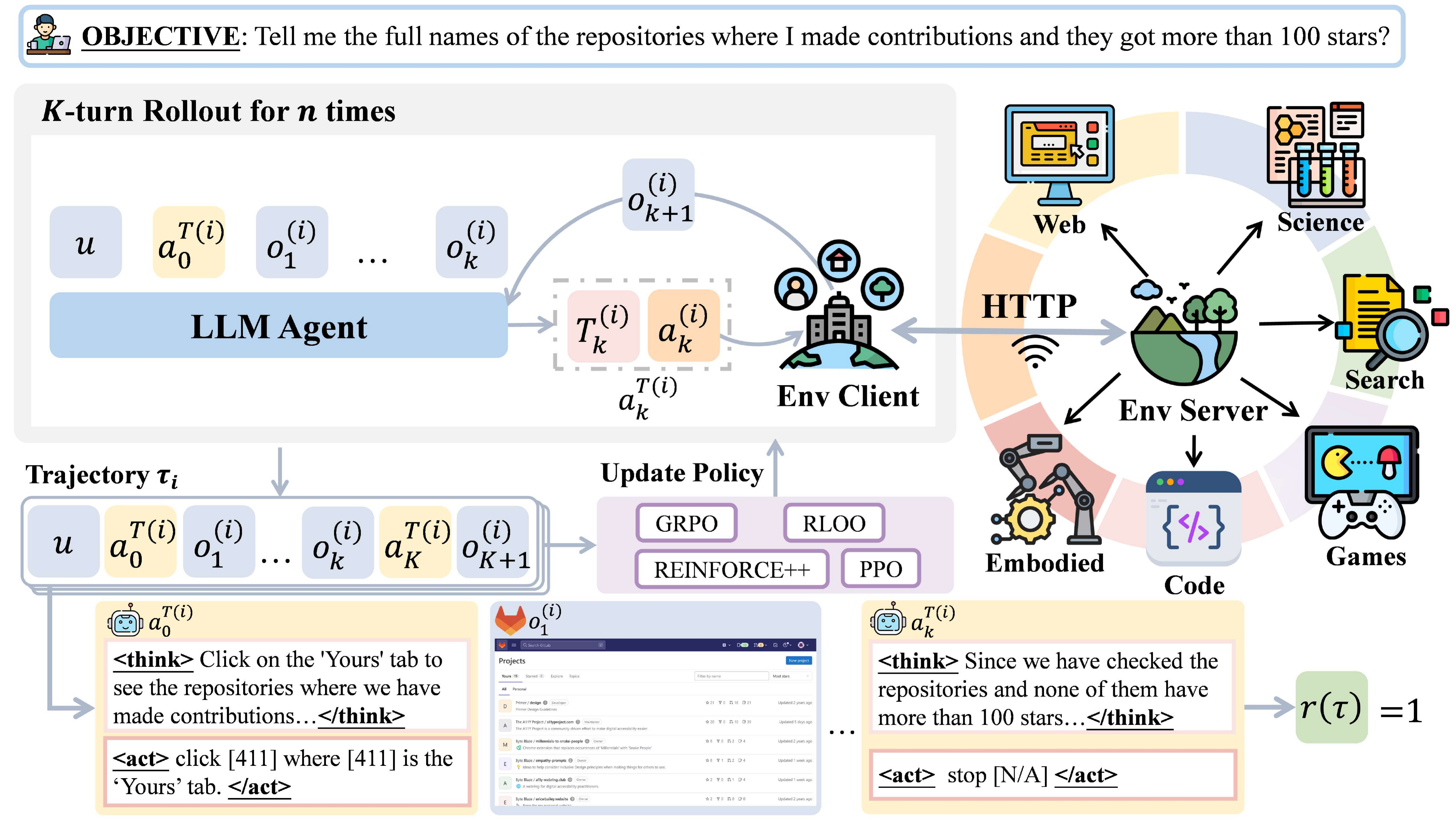}
    \caption{
    Overview of the AgentGym-RL framework. It features a decoupled, flexible, and extensible architecture, comprising three primary modules—the environment, the agent, and the training module. It supports diverse scenarios, environments, and algorithms. 
    }
    \label{fig:AgentGym-main}
\end{figure*}

To bridge this gap, we introduce AgentGym-RL, a new framework for training LLM agents for multi-turn interactive decision-making through RL (Figure \ref{fig:AgentGym-main}). Designed with a modular and decoupled architecture, AgentGym-RL enables clean separation of agents, environments, and learning algorithms—offering high extensibility and flexibility for diverse research needs. The framework supports mainstream RL algorithms, including PPO \citep{PPO}, GRPO \citep{GRPO}, and REINFORCE++ \citep{REINFORCEpp}, and is equipped with a wide range of real-world scenarios, e.g., web navigation \citep{WebArena, Mind2Web, WebShop}, deep search \citep{BrowseComp, Search-R1}, digital games \citep{ADaPT, MineDojo}, embodied tasks \citep{BabyAI, ALFWorld}, and scientific tasks \citep{ScienceWorld, PaperBench}.

Furthermore, to tackle the exploration–exploitation trade-off and improve optimization stability in agent RL training, we propose ScalingInter-RL, a method that progressively extends the agent–environment interaction horizon during training.
The core insight of this approach is to let the agent adapt to the environment in stages: beginning with exploitation to achieve reliable mastery of basic skills and simple tasks; subsequently increasing interaction horizon to promote exploration, refine behaviors, overcome shortcuts, and address more complex challenges. 
This progressive interaction-scaling strategy enables the agent to uncover richer interaction patterns (e.g., planning and reflection) and cultivate a broader set of skills and behaviors over time.

Our extensive experiments prove that AgentGym-RL delivers consistent and significant performance gains for agents across five tasks spanning 5 scenarios (Figure \ref{fig:main_greedy_performance}). Open-source models (e.g., Qwen-2.5-7B \citep{qwen_2_5}) trained with our framework and method achieved an average improvement of 33.65 points, matching—or even outperforming—larger commercial, closed-source models such as OpenAI-o3 \citep{openai202x_o3o4mini} and Gemini-2.5-Pro \citep{comanici2025gemini}. We also conducted numerous analytical experiments to provide key findings and insights, showing that scaling post-training and test-time compute have significant potential for developing agentic intelligence (Figure \ref{fig:model_performance_scatter}). We hope our work will be a valuable contribution to the community’s progress.

In summary, our main contributions are:
\begin{enumerate}
    \item We propose and open-source AgentGym-RL, a new unified, modular, and flexible end-to-end RL framework designed for agent multi-turn interactive decision-making that includes a diversity of scenarios and environments.
    \item We propose ScalingInter-RL, a progressive interaction-scaling framework that incrementally adapts agents to their environment, facilitating the refinement of interaction patterns and skill acquisition. It enhances optimization stability in RL and achieves a balance between exploration and exploitation.
    \item Our extensive experiments demonstrate that AgentGym-RL and ScalingInter-RL deliver significant and consistent performance gains, matching or exceeding commercial models. In addition, we conduct empirical analyses that yield critical insights into agent design and operational paradigms, offering valuable guidance and resources for future research.
\end{enumerate}

\section{Preliminaries}
\subsection{Formulation}
In this work, we study the multi-turn interactive decision-making tasks, i.e., agentic tasks, and we model them as a Partially Observable Markov Decision Process (POMDP) $(\mathcal{U, S, A, O, T}, r)$ like \citet{AgentGym, DBLP:conf/icml/ZhouZPLK24}, where $\mathcal{U, S, O, T: S \times A \to S}, r\mathcal{: U \times S \to \mathbb{R}}$ represents the instruction space, the state space, the action space, the observation space, the deterministic state transition function, and the reward function, respectively.

Given a task instruction $u \in \mathcal{U}$, the agentic task requires the LLM agent to generate a sequence of actions $a_k^{T} \sim \pi_\theta(\cdot | s_k)$ based on its policy $\pi_\theta$ parameterized by $\theta$ to complete the given task, where $a_k \in \mathcal{A}$, and $s_k \in \mathcal{S}$, and $T$ is the thinking path \citep{ReAct}. The agent then receives an observation $o_k \in O$ from the environment, and the state is then transitioned to $\mathcal{T}(s_k, a_k) = s_{k+1}$. Finally after $N$ turns of interactions, the environment $e$ provides an outcome reward $r(\tau) \in [0, 1]$ to describe the completion of the multi-turn interactive decision-making tasks.

\subsection{Policy Gradient}
We utilize policy gradient methods \citep{PGSutton} that optimizes our policy to maximize the expected cumulative reward. Unlike value-based methods that estimate the value function to derive a policy, policy gradient methods directly search the policy parameter space to find the optimal policy.

The core idea of policy gradient methods is to perform gradient ascent according to the objective $J(\theta)$, which is a function of the policy parameters $\theta$. Specifically, $J(\theta)$ represents the expected cumulative reward the agent anticipates receiving when following policy $\pi_\theta$ and interacting with the environment. Mathematically, this is expressed as the expectation of the total reward $r(\tau)$ over trajectories $\tau$ generated by the policy:

\begin{equation}
    J(\theta) = \mathbb{E}_{\tau \sim \pi_\theta} \left[ r(\tau) \right]
\end{equation}

To perform gradient ascent on $J(\theta)$, we require the policy gradient $\nabla_\theta J(\theta)$. In the vanilla policy gradient methods, the policy gradient can be estimated by:

\begin{equation}
    \nabla_\theta J(\theta) = \mathbb{E}_{\tau \sim \pi_\theta} \left[ r(\tau) \sum_{k=0}^{K} \nabla_\theta \log \pi_\theta(a_k | s_k) \right]
\end{equation}

where $\pi_\theta$ is the policy parameterized by $\theta$, $\tau$ represents a trajectory consisting of a sequence of states and actions, $a_k$ and $s_k$ are the action and state at time step $k$, and $r(\tau)$ is the reward of the trajectory $\tau$.

With the policy gradient estimated, we can optimize the parameters $\theta$ of the policy $\pi_\theta$ towards a direction of maximizing the expected cumulative reward with the gradient descent method by:

\begin{equation}
    \theta_{new} = \theta_{old} - \alpha \nabla_\theta J(\theta)
\end{equation}

where $\alpha \in [0, 1]$ is the learning rate. 
Mainstream RL algorithms for training LLMs include PPO \citep{PPO}, GRPO \citep{GRPO}, REINFORCE++ \citep{REINFORCEpp}, and RLOO \citep{RLOO}—all of which are integrated into our framework.

\section{The AgentGym-RL Framework}
\subsection{Architecture Overview}
The AgentGym-RL framework is built on AgentGym \citep{AgentGym}, which provides several basic interactive environments for LLM agents. Our main extensions focus on three aspects: 
\begin{enumerate}
    \item Introducing more realistic environments and tasks (e.g., Deep Search tasks) to facilitate the development of more general agents.
    \item Incorporating a diverse set of online reinforcement learning algorithms covering both classical and state-of-the-art methods, to ensure consistency with current research frontiers, and offer an extensible foundation for the community to build upon.
    \item Implementing extensive engineering optimizations and agent–environment co-design, such as improved rollout parallelization and memory-leak mitigation.
\end{enumerate}

As shown in Figure~\ref{fig:AgentGym-main}, the framework is organized into three main components: 
\begin{itemize}
    \item The \textbf{Environment} module provides diverse scenarios via a standardized server–client architecture with unified HTTP protocols. 
    \item The \textbf{Agent} module encapsulates the reasoning and decision-making process of agents in multi-turn interactions, with support for advanced mechanisms such as long-horizon planning and self-reflection. 
    \item The \textbf{Training} module implements reinforcement learning pipelines and other training methods to optimize agent policies. 
\end{itemize}
A more detailed description of our architecture is shown in Appendix \ref{appendix:architecture}.

\begin{figure*}[t]
    \centering
    \includegraphics[width=0.99\linewidth]{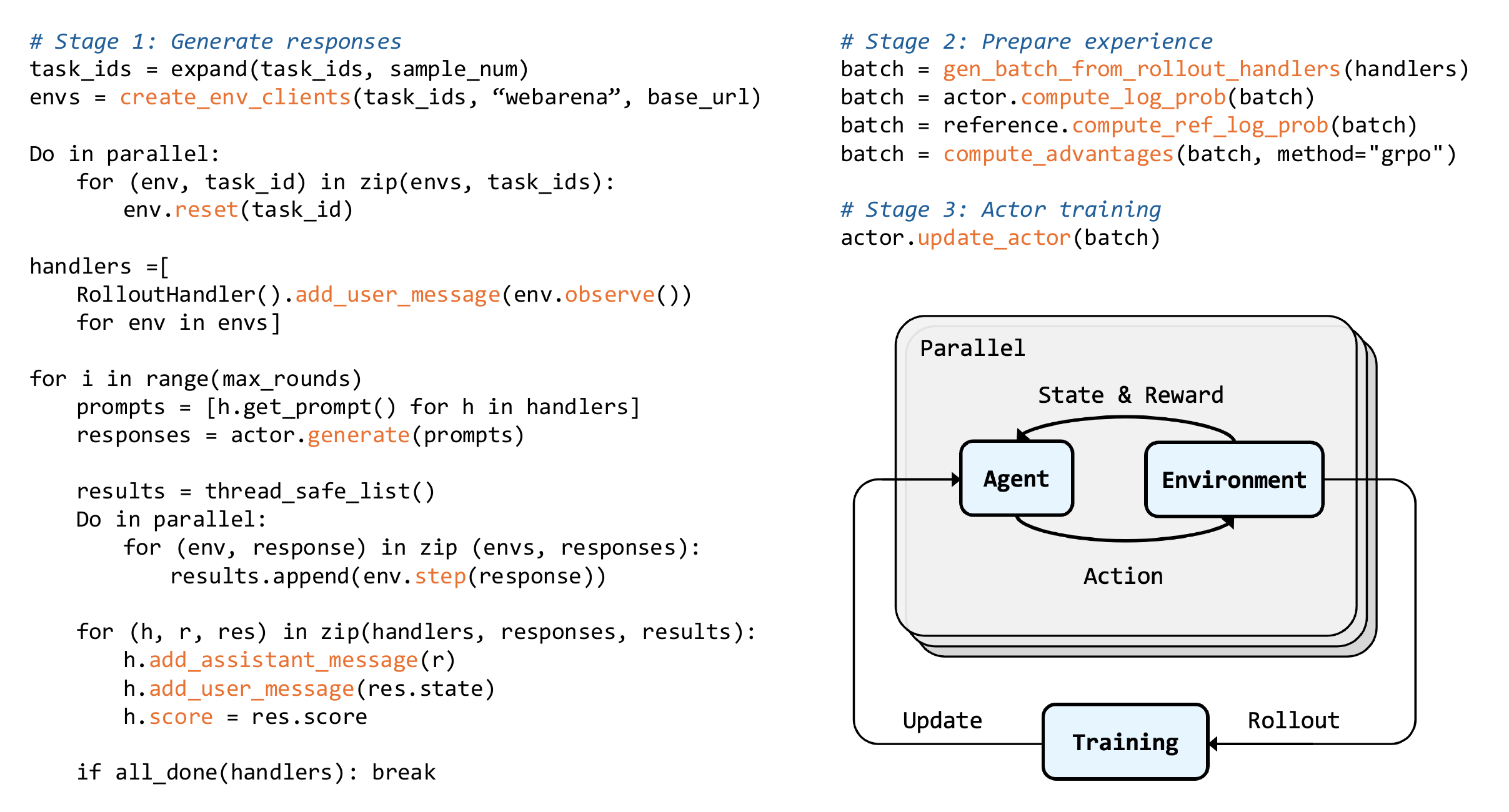}
    \caption{
    Pseudocode demonstrating the example usage of our proposed framework (provided APIs marked orange), alongside a simplified theoretical diagram illustrating the agent - environment interaction and training pipeline.
    }
    \label{fig:pseudocode}
\end{figure*}

Given a batch of user queries and initial environment states, our framework initializes multiple independent environment clients in parallel. Each client interacts exclusively with a single agent, ensuring that executions are isolated and non-interfering. In every client, the agent generates an action that is executed in the environment, which then returns the updated state and reward for the next decision. A batch of such trajectories is collected concurrently across clients and subsequently fed into the training module to update the agent policy. The overall workflow and corresponding pseudocode of our framework are shown in Figure \ref{fig:pseudocode}.

\subsection{Features and Characteristics}
In this section, we highlight the key features of the AgentGym-RL framework, covering four aspects: environment coverage, algorithm support, architectural advantages, and open-source contributions.

\subsubsection{Diverse scenarios and environments.}

To build LLM agents capable of multi-turn sequential decision-making for complex tasks in real-world environments, AgentGym-RL covers a broad spectrum of scenarios to comprehensively evaluate and foster the agent's ability to perceive its environment, long-term planning towards a goal, in-depth reasoning for making intelligent decisions, aptitude for reflection and correction when facing setbacks or making mistakes.

It includes:
\begin{itemize}
    \item \textbf{Web Navigation}: Interacting with dynamic websites for tasks such as booking flights or extracting structured information, which requires agents to follow instructions, interpret textual and visual content, manipulate dynamic interfaces, and plan multi-step actions.  
    \item \textbf{Deep Search}: Performing multi-step, goal-directed queries with tools like browsers or Python interpreters, demanding strong information-seeking, multi-hop reasoning, long-term memory, and knowledge synthesis across sources. 
    \item \textbf{Digital Games}: Exploring and solving problems in interactive game-like environments, emphasizing real-time decision-making, strategy development, and adaptability to complex, dynamic settings. 
    \item \textbf{Embodied Tasks}: Controlling virtual or physical bodies for navigation, manipulation, and task execution, which calls for goal-directed planning, spatial reasoning, and robust perception–action grounding. 
    \item \textbf{Scientific Tasks}: Conducting experiments and solving problems in physically grounded, knowledge-intensive settings, requiring precise execution, dynamic interpretation of feedback, evidence-based reasoning, and iterative hypothesis refinement. 
\end{itemize}

\subsubsection{Comprehensive RL algorithm support.} 

While the original AgentGym supports only a limited set of training methods based on supervised fine-tuning, AgentGym-RL places online reinforcement learning at its core, empowering agents to dynamically explore and adapt through continuous interactions with the environment.

AgentGym-RL implements a suite of mainstream online RL algorithms:  
(1) \textbf{PPO}\citep{PPO}, a policy gradient method that improves training stability by clipping policy updates to prevent overly large steps, simplifying the trust-region concept from TRPO\citep{TRPO} while maintaining strong empirical performance;  
(2) \textbf{GRPO}\citep{GRPO}, a PPO-derived method that normalizes rewards within groups of sampled actions per state and applies PPO-style clipping, reinforcing higher-performing actions relative to others;  
(3) \textbf{RLOO}\citep{RLOO}, a REINFORCE variant that uses the average reward of other samples in the same batch as a per-sample baseline, reducing variance in policy gradient estimates;  
(4) \textbf{REINFORCE++}\citep{REINFORCEpp}, an enhanced REINFORCE\citep{REINFORCE} algorithm that integrates PPO-style clipping and KL penalties, enabling more stable, simpler, and computationally efficient training without the need for a critic network. 

Beyond online RL, the framework also supports a broad range of complementary training paradigms:  
\textbf{SFT (Supervised Fine-Tuning)}\citep{SFT} is a standard training method where the agent learns to imitate expert demonstrations or golden trajectories step by step.
\textbf{DPO (Direct Preference Optimization)}\citep{DPO} is a variant of reinforcement learning that does not involve online interaction with the environment; instead, it learns from pre-collected preference pairs.
For rejection sampling\citep{rejection_sampling}, we support methods like \textbf{AgentEvol}\citep{AgentGym}, which iteratively fine-tunes agents on trajectories generated by themselves and filtered based on task success.

\subsubsection{Extensibility, Scalability, and Reliability.}

Since AgentGym-RL is primarily designed to support large-scale reinforcement learning research and development for the community, we have carried out extensive engineering design, practice, and optimization to ensure the framework’s extensibility, scalability, and reliability.

\begin{itemize}
    \item \textbf{Extensibility}. 
    Extensibility is critical for supporting evolving research needs, allowing a framework to accommodate new environments, agent architectures, and training strategies without disrupting existing components. In our system, we adopt a modular and decoupled design, where the core components—Environment, Agent, and Training—are fully plug-and-play. As a result, researchers can easily incorporate new reinforcement learning objectives, reward functions, or sampling techniques, facilitating reproducible experiments and enabling exploration across a wide spectrum of algorithmic directions. For example, a new environment can be introduced by simply inheriting from \texttt{BaseEnvClient} and implementing the required methods such as \texttt{reset()}, \texttt{step()}, and \texttt{observe()}. Once implemented, the new environment can be seamlessly used with existing agent architectures and training routines, enabling rapid experimentation without modifying any of the core framework components.

    \item \textbf{Scalability}. Recent advances in reinforcement learning increasingly rely on large-scale training, involving massive amounts of data and extended interaction sequences, which poses significant challenges for system scalability. To meet these challenges, a framework must be able to scale both in parallelism and interaction duration. We implemented a series of optimizations to achieve this. For example, we replaced WebArena’s default single-browser-per-process design with a subprocess-based architecture, enabling a single server to manage multiple Chromium instances concurrently and thereby enhancing parallelism. Similarly, in SciWorld environment, we redesigned the environment’s initialization and reset routines to support robust parallel creation and resetting of multiple instances, resolving previous failures in concurrent instantiation. In addition, we support longer training horizons through a \texttt{full-reset} interface in WebArena, which restores each web server to its initial state after every episode and mitigates state inconsistencies over time. Together, these optimizations allow our framework to scale effectively, facilitating large-scale training and enabling the research community to conduct a broad range of experiments.

    \item \textbf{Reliability}. Large scale multi-turn agent RL training poses a significant challenge to system reliability, that is, the ability to maintain consistent and reliable operation over long training periods. To achieve reliability, a framework must prevent failures that could disrupt training and ensure that critical resources are managed correctly. For instance, we optimized the memory management implementation in TextCraft. The original environment suffered from a memory leak in its recursive \texttt{crafting\_tree} implementation, where redundant self-replication of a list structure caused exponential memory growth and eventual crashes during training. We resolved this issue by refactoring the recursion to eliminate redundant copies. Likewise, in SciWorld, a memory leak in its internal clock mechanism caused progressive memory accumulation and instability during extended rollouts. We addressed this issue by refactoring the clock implementation to eliminate the leakage. Through our optimization, the framework provides a reliable environment for long-horizon training, ensuring consistent and uninterrupted operation over extended interaction sequences.
\end{itemize}

Collectively, these design and optimizations remove major engineering bottlenecks and make reproducible, large-scale RL experiments feasible across heterogeneous environments. 

\begin{figure*}[t]
    \centering
    \includegraphics[width=\linewidth]{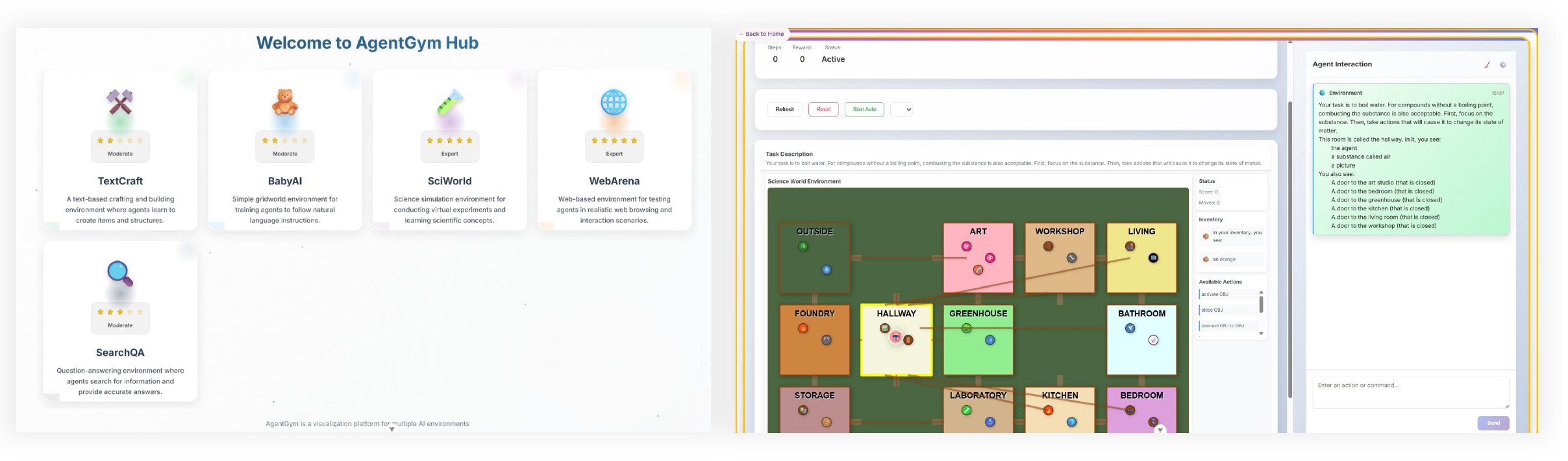}
    \caption{An overview of the visualized user interface of our framework.}
    \label{fig:env}
\end{figure*}

\subsubsection{Open-source availability and community extensibility.}

We design AgentGym-RL to foster a collaborative ecosystem where community contributions directly accelerate methodological progress while upholding verifiable research standards. AgentGym-RL is released as an open-source framework under permissive licensing, built upon established open-source frameworks such as veRL\citep{verl} and AgentGym\citep{AgentGym} while maintaining full open-source availability. The framework provides comprehensive documentation, reproducible training pipelines, and standardized APIs to ensure research transparency and practical adoption. Its modular architecture-which includes clearly defined extension points-enables the seamless integration of new environments and training methods, allowing the research community to extend functionality without disrupting the core workflows. To facilitate probing of data and model behaviors, we provide an interactive user interface, which streamlines empirical analysis for iterative development.

\paragraph{Usability, reproducibility and standardized evaluation.}
AgentGym-RL is designed to be user-friendly for the community. To systematically address reproducibility challenges in LLM-based reinforcement learning, AgentGym-RL institutes a standardized evaluation process and reproducible training pipelines. This design enforces uniform metrics and consistent experimental procedures to ensure fair comparisons. We provide easy-to-setup reproduction scripts that automate the entire workflow, from environment configuration to final evaluation. This design enables researchers to replicate prior findings with high fidelity and significantly lowers the barrier for building upon existing work, thereby promoting verifiable research standards.

\paragraph{Visualized user interface for observability and analysis.}

As shown in Figure~\ref{fig:env}, AgentGym-RL includes an interactive user interface designed to facilitate the probing of data and model behaviors. This tool streamlines empirical analysis by enabling researchers to perform a fine-grained, step-by-step inspection of an agent's decision-making process. It allows for the replay and examination of full interaction trajectories, visualizing the interplay between environmental observations, the agent's internal reasoning, and its resulting actions. This capability provides direct insights into model performance and failure modes, thereby accelerating the iterative development and debugging cycle.

\subsection{ScalingInter-RL: Progressive Scaling Interaction for Agent RL}

\begin{figure*}[t]
    \centering
    \includegraphics[width=0.99\linewidth]{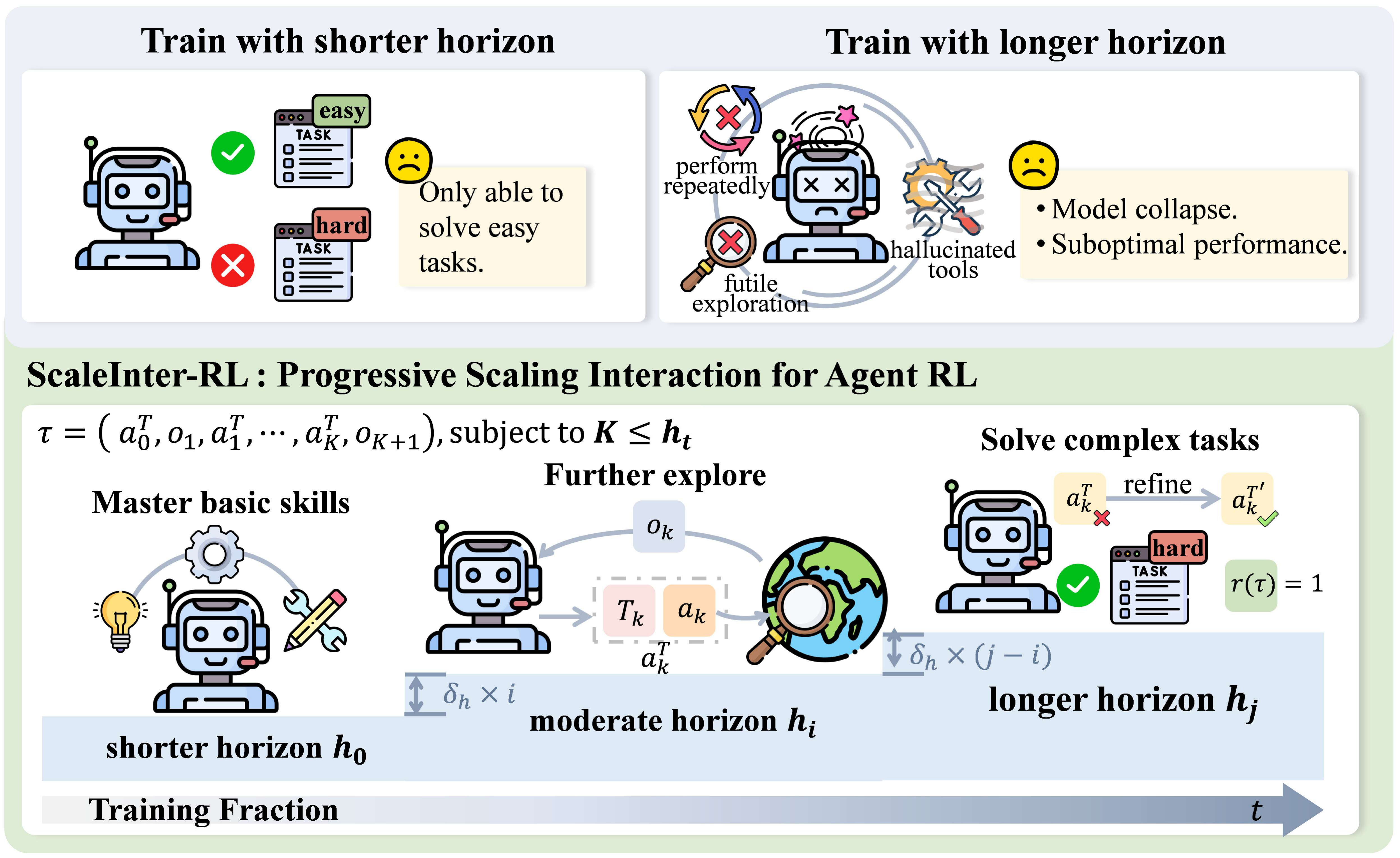}
    \caption{Illustration of the ScalingInter-RL approach. It allows the agent to adapt in stages: initially, by limiting interaction turns to prioritize exploitation, master basic skills, and solve easy tasks; later, by gradually increasing interactions to explore, avoid shortcuts, refine behavior, and tackle harder problems. Ultimately, this process trains a stronger agent.}
    \label{fig:ScalingInter-RL-Method}
\end{figure*}
\paragraph{Motivation and core insight.}

When assigned a task, an agent engages in iterative interactions with the environment—observing changes, reasoning about them, and executing subsequent actions. Through this cycle, the agent explores and experiments thoroughly, ultimately reaching the target state. This process is analogous to inference–compute scaling in LLM reasoning (as exemplified by OpenAI o1 and DeepSeek-R1), where additional computational resources are allocated at test time or during RL rollouts, allowing the model to reason more deeply before producing a final answer.

In comparison, we argue that beyond relying on internal reasoning to select the next action, agents should also expand their external interactions with the environment to ensure sufficient exploration and accumulate richer context toward the final goal—capturing a form of practice-driven insight. Yet, our preliminary experiments indicate that beginning with a large number of interaction turns often leads the model into redundant reasoning and unproductive actions, ultimately causing training collapse and degraded performance. Conversely, constraining the number of interactions to remain consistently small tends to narrow exploration and limits the agent’s ability to master diverse patterns. This motivates us to propose our method.

\paragraph{Method.} 
We draw inspiration from reinforcement learning for LLM reasoning \citep{DeepSeek-R1, openai_o1, r3_2024, ReFT, kimi_k1.5, Skywork} and propose ScalingInter-RL, a training approach designed to balance exploration and exploitation while ensuring stable optimization. At its core is a progressive horizon-scaling strategy that adaptively adjusts the number of interaction turns during RL. The objective is to maximize the expected terminal reward under a constrained interaction budget:
\[
J(\theta) = \mathbb{E}_{\tau \sim \pi_\theta}\left[ r\left(\tau\right) \right],
\]

where each trajectory $\tau = \left( a_0^{T}, o_1, a_1^{T}, \dots, a_{K-1}^{T}, o_{K} \right)$ is sampled from the current policy $\pi_\theta$, with $K$ representing the total number of interaction turns. To enable the agent to rapidly learn effective behaviors under limited interaction resources, we begin training with a small horizon. By initially constraining the horizon, the agent learns to exploit its policy with maximum efficiency, achieving early proficiency on simple tasks, and laying the groundwork for deeper, long-horizon reasoning. As training progresses, we introduce a monotonic schedule $\{h_1 < h_2 < \cdots < h_n\}$, where $h_t$ defines the maximum number of interaction turns allowed during phase t:

\[
\tau_{t} \sim \pi_\theta\left(\tau \mid h_t\right), \quad \text{subject to } K_t \leq h_t.
\]

The horizon $h_t$ is updated every $\Delta$ training steps according to a curriculum schedule:

\[
h_{t+1} = h_t + \delta_h,
\]

where $\delta_h$ is an adaptive increment. As the horizon increases, the agent is incentivized to explore longer decision paths, facilitating the emergence of higher-order cognitive behaviors such as planning, reflection, and strategic backtracking, which is similar to the length-scaling phenomenon in RLVR for large reasoning models~\citep{DeepSeek-R1,cao2025skyrl,deepscaler2025}. This phased scaling allows ScalingInter-RL to align the depth of interaction with the agent’s evolving policy capabilities, bridging efficient early-stage exploitation and long-horizon generalization.

\section{Experiments}

To verify the stability and effectiveness of the AgentGym-RL framework, we conduct extensive experiments across a diverse set of scenarios and environments. Our results demonstrate that LLM agents are capable of exploring and learning from scratch based solely on environment feedback, without the need for prior supervised fine-tuning, ultimately achieving performance that is comparable to, or even surpasses, that of commercial closed-source models such as OpenAI o3.

\subsection{Experimental Settings}
\paragraph{Scenarios, Environments and Tasks.}
As mentioned before, we include five scenarios in AgentGym-RL.
Specifically, for web navigation, we include WebArena \citep{WebArena} which is a realistic and reproducible web environment containing four distinct domains prevalent on the internet: online shopping, discussion forums, collaborative development, and business content management; for deep search, we include a RAG-based environment \citep{Search-R1, TriviaQA, 2WikiMultiHopQA, NQ, PopQA, Musique, HotpotQA, Bamboogle} which enables LLMs to interact with search engines and solve multi-turn retrieval and reasoning tasks; for digital games, we include TextCraft \citep{ADaPT}, a text-based crafting game environment in which agents complete tasks via natural language interactions and task-based planning; for embodied tasks, we include BabyAI \citep{BabyAI} which provides a controllable grid world with text instructions for embodied reasoning in simulated environments; for scientific tasks, we include SciWorld \citep{ScienceWorld} which offers a scientific exploration simulator where agents conduct scientific experiments through text-driven reasoning cycles.

\paragraph{Baselines and backbone models.}

We leverage Qwen-2.5-3B and Qwen-2.5-7B \citep{qwen_2_5} as our primary backbone models. We introduce the closed-source Gemini 2.5 Pro \citep{comanici2025gemini}, OpenAI o3 \citep{openai202x_o3o4mini}, and GPT-4o \citep{gpt4o} as baselines. Additionally, we include the open-source DeepSeek-R1 \citep{DeepSeek-R1}, Qwen-2.5-72B \citep{qwen_2_5}, Llama-3.1-8B \citep{llama3}, and Llama-3.1-70B \citep{llama3} models for comparison.

\paragraph{Detailed settings of each environment.} 
We provide detailed descriptions of the tools, APIs, and experimental settings for each environment in Appendix  \ref{appendix: Detailed Settings of Each Environment}.

\begin{figure*}[ht]
    \centering
    \includegraphics[width=0.99\linewidth]{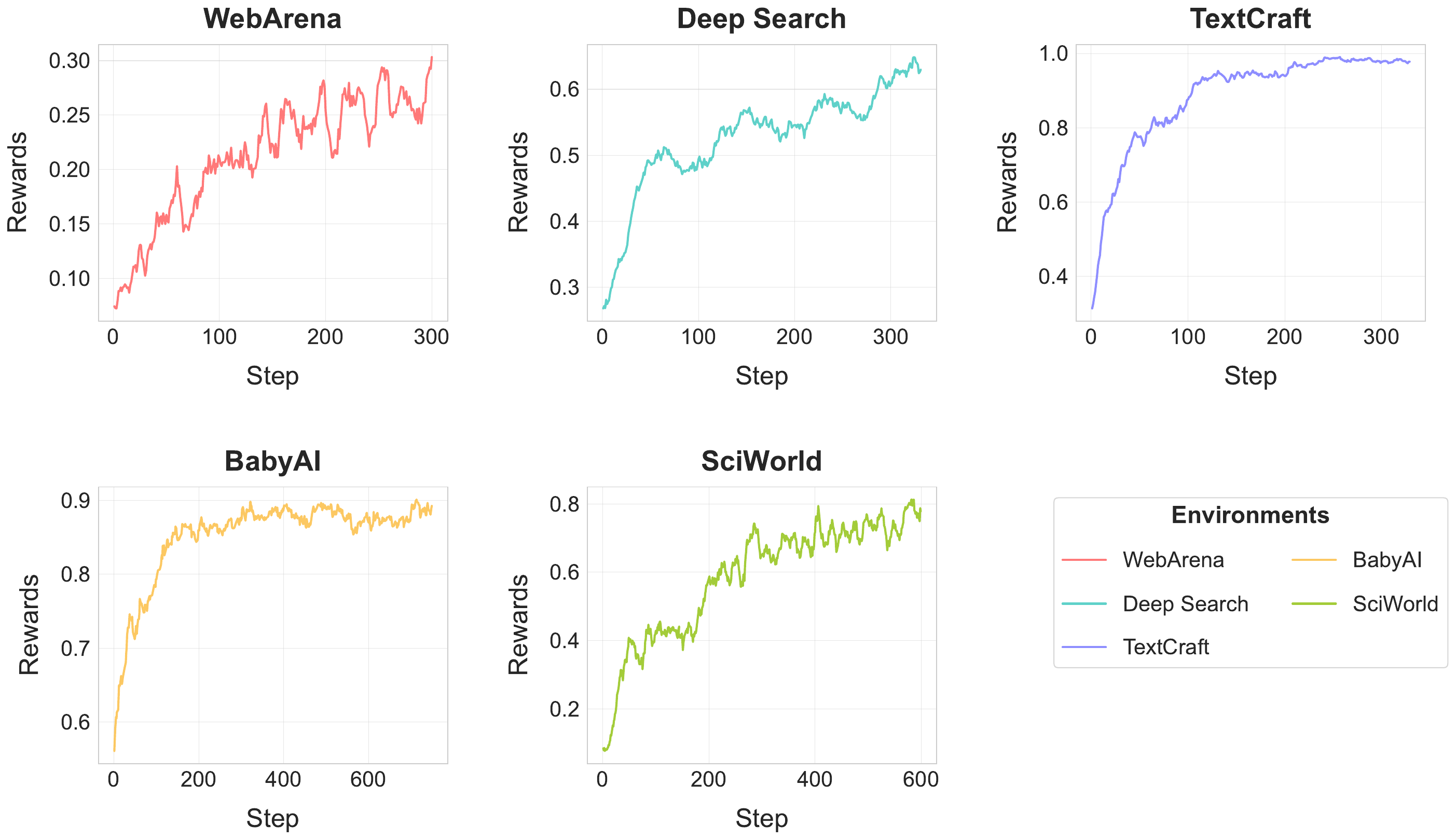}
    \caption{Training rewards in different environments.}
    \label{fig:environment_rewards}
\end{figure*}

\subsection{Overall Results, Findings, and Insights}
The main results are shown in Figure~\ref{fig:combined_performance}, Table~\ref{tab:webarena_results}, Table~\ref{tab:searchqa_results}, Table~\ref{tab:textcraft_results}, Table~\ref{tab:babyai_results}, and Table~\ref{tab:sciworld_results}. In this section, we discuss the overall findings and insights.

\paragraph{Reinforcement learning generally improves agentic intelligence of open-source LLMs to the level of proprietary models.} 
As illustrated in Figure~\ref{fig:combined_performance}, the AgentGym-RL-7B model not only outperforms other open-source models by a large margin but also demonstrates a clear lead in average success rate over leading closed-source models like GPT-4o and Gemini-2.5-Pro across five different scenarios. This achievement highlights our framework's effectiveness in enabling models to learn and make decisions in complex interactive tasks, successfully bridging the performance gap between open-source and proprietary models on advanced intelligent assignments.

\paragraph{ScalingInter-RL boosts performance significantly and consistently.}
Instead of relying on extensive hyperparameter tuning, we set the transition points between phases according to the total optimization steps of the original RL process. As shown in our results, ScalingInter-RL consistently outperforms the baseline across diverse environments and tasks. Notably, it delivers more than a $10$\% improvement on WebArena, bringing performance close to that of closed-source commercial models. On the TextCraft benchmark, it surpasses the base model by $30$ points, achieving state-of-the-art results. These findings highlight the effectiveness of our approach in striking a balance between exploration and exploitation in reinforcement learning. As illustrated in Figure \ref{fig:environment_rewards}, experiments across different environments show that leveraging our AgentGym-RL framework with the ScalingInter-RL algorithm yields stable, sustained, and substantial reward improvements.

\begin{figure*}[t]
    \centering
    \includegraphics[width=\linewidth]{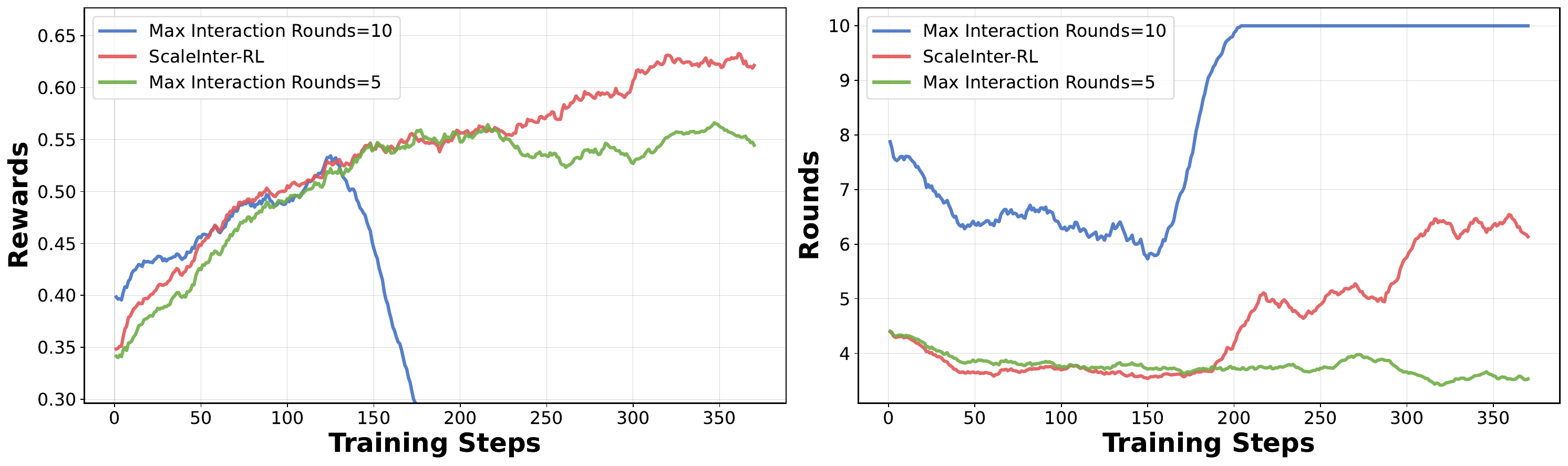}
    \caption{Training dynamics under different maximum interaction turns in Deep Search environment. Longer-turn settings (e.g., 10) initially achieve higher rewards by enabling richer exploration, but rapidly collapse due to high variance, credit assignment difficulties, and overfitting to spurious behaviors. Shorter turns (e.g., 5) yield more stable but less exploratory learning, leading to a performance ceiling. Our ScalingInter-RL method progressively increases the interaction horizon, and ultimately achieves higher and more efficient long-term performance.}
    \label{fig:searchqa_training_dynamics}
\end{figure*}

\paragraph{Large interaction budget accelerates early gains but ultimately leads to unstable training.}
As shown in Figure \ref{fig:searchqa_training_dynamics}, we observe that using a larger maximum interaction turn (e.g., 10) achieves higher performance in the early stage compared to a shorter-turn setting (e.g., 5), but rapidly collapses as training progresses. 
This indicates that excessive exploration in early stages of training is not necessarily a good choice. Before establishing a solid foundation, the agent may perform unproductive and inefficient exploration, leading to the risk of training instability. 
By contrast, shorter rounds restrict early exploration but provide more stable learning signals, leading to more reliable long-term performance. Taken together, these contrasting dynamics between longer and shorter turns motivate our ScalingInter-RL method, which progressively extends the interaction horizon during training.

\paragraph{ScalingInter-RL demonstrates more stable and efficient training dynamics during RL optimization.}
As shown in Figure \ref{fig:searchqa_training_dynamics}, our method is initially constrained by the number of interaction turns. Although it struggles to fully master difficult tasks at first, by exploiting foundational skills and knowledge it achieves a noticeable increase in rewards; later, as it engages in more turns of interaction and exploration with the environment, it shapes its reasoning paradigm and interaction behaviors, ultimately reaching a high level of performance. In contrast, RL with fewer turns yields diminishing returns in later stages and hits a performance ceiling, while RL with large interaction budget quickly collapses.
Furthermore, just as is observed with RL for reasoning models~\citep{cao2025skyrl}, our gradual scaling of interactions dramatically reduces the computational resources and time required in RL phase, enabling more efficient optimization.

\paragraph{Post-training and test-time compute show higher scaling potential than model size.} 
A key insight from our experiments is that strategic investment in post-training and test-time computation is more impactful than merely increasing a model's parameter count. Figure \ref{fig:combined_performance} (right) clearly illustrates this point: our ScalingInter-RL model, with only 7B parameters, achieves an average success rate of approximately 58.6\% after being trained with our reinforcement learning framework. This performance not only surpasses other open-source models of similar size but also significantly outperforms much larger models like Llama3.1-70B (~47\%) and Qwen2.5-72B (~43\%), which have nearly ten times the parameters. This demonstrates that the performance improvement gained from simply scaling model size is limited and less steep compared to the gains from targeted post-training and inference-time computation using frameworks like AgentGym-RL.

\paragraph{Environmental structure is a key determinant for the efficiency of reinforcement learning.} 
The effectiveness of AgentGym-RL varies depending on the nature of the environment and feedback. In simulated worlds with clear rules and explicit cause-and-effect, like TextCraft, BabyAI, and SciWorld, RL delivers the most significant performance leaps. On SciWorld's complex scientific reasoning tasks, our method boosted the model's score from 1.50 to 50.50, an astounding increase of nearly 50 points. In contrast, for more open-ended environments like WebArena and Deep Search, the performance gains from RL were rather moderate, though still positive. In these tasks, agents must navigate the complexities of real websites, handle multi-step crafting plans, or process noisy information from search engines, making it more challenging to learn optimal strategies through trial and error. This suggests that while RL has broad applicability, it excels most in environments where clear feedback and successful pathways can be readily discovered through exploration.

\subsection{Detailed Task Performance across Environments}

\paragraph{Web navigation.}
\begin{table*}[!t]
\centering
\caption{Evaluation results on WebArena benchmark. For each group, the best result is in \textbf{bold}, and the second-best is \underline{underlined}. In the first row, G \& R means GitLab and Reddit.}
\label{tab:webarena_results}
\resizebox{0.6\linewidth}{!}{%
\begin{tabular}{lccccc}
\toprule
\multicolumn{1}{c}{\textbf{Model}} & \multicolumn{1}{c}{\textbf{Shopping}} & \multicolumn{1}{c}{\textbf{CMS}} & \multicolumn{1}{c}{\textbf{Maps}} & \multicolumn{1}{c}{\textbf{G \& R}} & \multicolumn{1}{c}{\textbf{Overall}} \\

\midrule

%--------------------------------- Proprietary LLMs ---------------------------------%
\rowcolor{gray!10}\multicolumn{6}{c}{\emph{Proprietary Models}} \\
\texttt{GPT-4o} & 20.00 & 13.33 & 10.00 & 20.00 & 16.00 \\
\texttt{Qwen-Max} & 20.00 & 13.33 & 20.00 & 30.00 & 20.00 \\

\texttt{Gemini-2.5-Flash} & \underline{26.67} & \underline{20.00} & 10.00 & 30.00 & 22.00 \\
\texttt{OpenAI o4-mini} & \textbf{33.33} & \textbf{26.67} & 20.00 & \underline{70.00} & \textbf{36.00} \\
\texttt{OpenAI o3} & \textbf{33.33} & 0.00 & \textbf{40.00} & \textbf{80.00} & \underline{34.00} \\

\texttt{Gemini-2.5-Pro} & \underline{26.67} & \textbf{26.67} & 0.00 & 60.00 & 28.00 \\

\midrule

%--------------------------------- Open-sourced LLMs ---------------------------------%
\rowcolor{gray!10}\multicolumn{6}{c}{\emph{Open-sourced Models $\geq$ 100B}} \\
\texttt{Qwen3-235B-A22B} & \underline{20.00} & \textbf{20.00} & \underline{20.00} & 20.00 & \underline{20.00} \\

\texttt{DeepSeek-V3-0324} & \underline{20.00} & \underline{13.33} & 10.00 & \underline{30.00} & 18.00 \\
\texttt{DeepSeek-R1-0528} & \textbf{33.33} & 6.67 & \textbf{30.00} & \textbf{50.00} & \textbf{28.00} \\
\midrule

%--------------------------------- Open-sourced LLMs ---------------------------------%
\rowcolor{gray!10}\multicolumn{6}{c}{\emph{Open-sourced Models $<$ 100B}} \\
\texttt{Qwen2.5-3B-Instruct} & 13.33 & 6.67 & \underline{10.00} & 10.00 & 10.00 \\
\texttt{Qwen2.5-7B-Instruct} & 14.29 & 6.67 & 0.00 & 16.67 & 9.76 \\
\texttt{Qwen2.5-72B-Instruct} & 13.33 & {13.33} & 0.00 & \underline{20.00} & 12.00 \\
\texttt{Qwen3-4B} & 13.33 & 6.67 & \underline{10.00} & 20.00 & 12.00 \\
\texttt{Qwen3-8B} & 20.00 & {20.00} & 0.00 & 10.00 & 14.00 \\
\texttt{Qwen3-32B} & {20.00} & 6.67 & \textbf{20.00} & 0.00 & 12.00 \\
\texttt{Llama-3.1-8B-Instruct} & 13.33 & 0.00 & \textbf{20.00} & \textbf{30.00} & {14.00} \\
\texttt{Llama-3.1-70B-Instruct} & \underline{26.67} & 6.67 & \textbf{20.00} & 10.00 & {16.00} \\

%--------------------------------- RL Models ---------------------------------%
\rowcolor{gray!10}\multicolumn{6}{c}{\emph{Our RL Models}} \\
\texttt{AgentGym-RL-3B} & {20.00} & \underline{26.67} & 10.00 & 10.00 & 18.00 \\

\texttt{AgentGym-RL-7B} & {20.00} & \textbf{33.33} & {0.00} & \textbf{30.00} & \underline{22.00} \\

\texttt{ScalingInter-7B} & \textbf{33.33} & \underline{26.67} & \textbf{20.00} & \underline{20.00} & \textbf{26.00} \\

\bottomrule
\end{tabular}%
}
\end{table*}
As shown in Table~\ref{tab:webarena_results}, our models demonstrate highly competitive performance on the WebArena benchmark. In particular, the ScalingInter-7B model achieves an overall accuracy of 26.00$\%$, significantly surpassing top-tier proprietary models like GPT-4o (16.00$\%$) and performing on par with larger models like DeepSeek-R1-0528 (28.00$\%$) and Gemini-2.5-Pro (28.00$\%$). Furthermore, another 7B model of ours, AgentGym-RL-7B, also achieved an overall score of 16.00$\%$, matching the performance of GPT-4o. This strong overall performance is underpinned by ScalingInter-7B's state-of-the-art proficiency in structured web navigation, where it achieved scores of 33.33$\%$ in Shopping and 26.67$\%$ in CMS, matching the best performance among all models in these categories. However, a significant performance gap remains when compared to the top-performing OpenAI o3 (34.00$\%$) and o4-mini (36.00$\%$), a disparity almost entirely concentrated in the "GitLab \& Reddit" sub-task.

\paragraph{Deep search.}
\begin{table*}[!t]
\centering
\caption{Evaluation results on Deep Search benchmark. For each group, the best result is in \textbf{bold}, and the second-best is \underline{underlined}. SearchR1-it-v0.3 baseline uses Search-R1-v0.3 models\citep{Search-R1-v0.3}}
\label{tab:searchqa_results}
\resizebox{0.9\textwidth}{!}{%
\begin{tabular}{lcccccccc}
\toprule
\multicolumn{1}{c}{\textbf{Model}} & \multicolumn{1}{c}{\textbf{NQ}} & \multicolumn{1}{c}{\textbf{TriviaQA}} & \multicolumn{1}{c}{\textbf{PopQA}} & \multicolumn{1}{c}{\textbf{HotpotQA}} & \multicolumn{1}{c}{\textbf{2Wiki}} & \multicolumn{1}{c}{\textbf{Musique}} & \multicolumn{1}{c}{\textbf{Bamboogle}} & \multicolumn{1}{c}{\textbf{Overall}} \\

\midrule

%--------------------------------- Proprietary LLMs ---------------------------------%
\rowcolor{gray!10}\multicolumn{9}{c}{\emph{Proprietary Models}} \\
\texttt{GPT-4o} & 20.00 & \textbf{70.00} & 30.00 & 30.00 & 32.00 & 10.00 & 34.00 & 26.75 \\
\texttt{Qwen-Max} & \underline{24.00} & 52.00 & 26.00 & 24.00 & 16.00 & 17.00 & 36.00 & 29.50 \\
\texttt{Gemini-2.5-Flash} & 8.00 & 60.00 & 30.00 & 24.00 & 16.00 & 8.00 & 34.00 & 23.50 \\
\texttt{OpenAI o4-mini} & 22.00 & \underline{68.00} & \underline{50.00} & 38.00 & 44.00 & \underline{28.00} & \underline{62.00} & \underline{42.50} \\
\texttt{OpenAI o3} & \textbf{28.00} & \textbf{70.00} & \textbf{56.00} & \textbf{46.00} & \textbf{64.00} & \textbf{29.00} & \textbf{74.00} & \textbf{49.50} \\

\texttt{Gemini-2.5-Pro} & 22.00 & 62.00 & 38.00 & 28.00 & 48.00 & 19.00 & 56.00 & 36.50 \\
\midrule

%--------------------------------- Open-sourced LLMs ---------------------------------%
\rowcolor{gray!10}\multicolumn{9}{c}{\emph{Open-sourced Models $\ge$ 100B }} \\

\texttt{Qwen3-235B-A22B} & \underline{28.00} & 54.00 & \underline{30.00} & \underline{32.00} & \underline{22.00} & \underline{14.00} & 32.00 & \underline{28.25} \\
\texttt{DeepSeek-V3-0324} & \underline{28.00} & \underline{60.00} & 24.00 & 28.00 & 18.00 & 11.00 & \underline{34.00} & 26.50 \\
\texttt{DeepSeek-R1-0528} & \textbf{32.00} & \textbf{68.00} & \textbf{42.00} & \textbf{44.00} & \textbf{50.00} & \textbf{21.00} & \textbf{44.00} & \textbf{40.25} \\
\midrule
%--------------------------------- Open-sourced LLMs ---------------------------------%
\rowcolor{gray!10}\multicolumn{9}{c}{\emph{Open-sourced Models $<$ 100B }} \\
\texttt{Qwen2.5-3B-Instruct} & 8.00 & 42.00 & 22.00 & 14.00 & 8.00 & 2.00 & 10.00 & 13.50 \\
\texttt{Qwen2.5-7B-Instruct} & 18.00 & 54.00 & 20.00 & 18.00 & 6.00 & 4.00 & 26.00 & 18.75 \\
\texttt{Qwen2.5-72B-Instruct} & 22.00 & 52.00 & 24.00 & 28.00 & 24.00 & 12.00 & \textbf{38.00} & 26.50 \\
\texttt{Qwen3-4B} & 18.00 & {58.00} & 26.00 & 24.00 & 26.00 & 5.00 & 20.00 & 22.75 \\
\texttt{Qwen3-8B} & 26.00 & 44.00 & 26.00 & 22.00 & 32.00 & 10.00 & 32.00 & 25.25 \\
\texttt{Qwen3-32B} & 24.00 & 54.00 & 22.00 & {36.00} & 28.00 & 11.00 & 20.00 & 25.75 \\

\texttt{Llama-3.1-8B-Instruct} & 16.00 & 26.00 & 12.00 & 6.00 & 2.00 & 4.00 & 18.00 & 11.00 \\
\texttt{Llama-3.1-70B-Instruct} & 20.00 & 44.00 & 22.00 & 22.00 & 18.00 & 9.00 & 32.00 & 22.00 \\

\texttt{SearchR1-it-3B-v0.3$_\text{GRPO}$} & 20.00 & 50.00 & {30.00} & 28.00 & 32.00 & 5.00 & 14.00 & 23.00 \\
\texttt{SearchR1-it-7B-v0.3$_\text{GRPO}$} & 24.00 & 52.00 & {30.00} & 22.00 & 34.00 & 6.00 & 26.00 & 25.00 \\

%--------------------------------- RL Models ---------------------------------%
\rowcolor{gray!10}\multicolumn{9}{c}{\emph{Our RL Models}} \\
\texttt{AgentGym-RL-3B} & {30.00} & 50.00 & {30.00} & 30.00 & \textbf{46.00} & 4.00 & 12.00 & 25.75 \\
\texttt{AgentGym-RL-7B} & \underline{44.00} & \underline{64.00} & \underline{32.00} & \underline{40.00} & {36.00} & \textbf{15.00} & \underline{26.00} & \underline{34.00} \\
\texttt{ScalingInter-7B} & \textbf{52.00} & \textbf{70.00} & \textbf{46.00} & \textbf{42.00} & \underline{44.00} & \underline{14.00} & {24.00} & \textbf{38.25} \\

\bottomrule
\end{tabular}%
}
\end{table*}
The evaluation results in Table~\ref{tab:searchqa_results} show the importance of sophisticated reasoning abilities, where proprietary models—particularly the OpenAI 'o' series—currently set the performance benchmark, with OpenAI o3 achieving the highest overall score of 49.50. Against this competitive landscape, our models demonstrate exceptional performance. Specifically, our ScalingInter-7B model achieved an excellent overall score of 38.25, not only surpassing top-tier proprietary models like GPT-4o (26.75) and Gemini-2.5-Pro (36.50) but also performing comparably to the strongest open-source model, DeepSeek-R1-0528 (40.25). Its strengths are particularly salient in key domains: it achieved the highest score overall on the NQ task (52.00) and tied for first place on TriviaQA (70.00) with GPT-4o. Furthermore, our AgentGym-RL-7B (34.00) and AgentGym-RL-3B (25.75) models also delivered strong results, each significantly outperforming open-source counterparts of similar or even larger scales. These results provide strong evidence that our reinforcement learning approach effectively unlocks the model's inherent reasoning capabilities, enabling it to reach or even exceed the performance of elite reasoning models in key scenarios--crucially, without the need for explicit additional long-reasoning.

\paragraph{Digital game.}
\begin{table*}[!t]
\centering
\caption{Evaluation results on TextCraft benchmark. For each group, the best result is in \textbf{bold}, and the second-best is \underline{underlined}.}
\label{tab:textcraft_results}
\resizebox{0.65\linewidth}{!}{%
\begin{tabular}{lccccc}
\toprule
\multicolumn{1}{c}{\textbf{Model}} & \multicolumn{1}{c}{\textbf{Depth 1}} & \multicolumn{1}{c}{\textbf{Depth 2}} & \multicolumn{1}{c}{\textbf{Depth 3}} & \multicolumn{1}{c}{\textbf{Depth 4}} & \multicolumn{1}{c}{\textbf{Overall}} \\

\midrule

%--------------------------------- Proprietary LLMs ---------------------------------%
\rowcolor{gray!10}\multicolumn{6}{c}{\emph{Proprietary Models}} \\
\texttt{GPT-4o} & \textbf{100.00} & 87.80 & \underline{64.00} & 0.00 & 83.00 \\
\texttt{Qwen-Max} & 93.55 & 75.61 & 36.00 & 0.00 & 69.00 \\
\texttt{Gemini-2.5-Flash} & \textbf{100.00} & \underline{95.12} & 40.00 & 0.00 & 80.00 \\
\texttt{OpenAI o4-mini} & \textbf{100.00} & \textbf{100.00} & \textbf{84.00} & 0.00 & \underline{93.00} \\
\texttt{OpenAI o3} & \textbf{100.00} & \textbf{100.00} & \textbf{84.00} & 0.00 & \underline{93.00} \\

\texttt{Gemini-2.5-Pro} & \textbf{100.00} & \textbf{100.00} & \textbf{84.00} & \textbf{33.33} & \textbf{94.00} \\
\midrule

%--------------------------------- Open-sourced LLMs ---------------------------------%
\rowcolor{gray!10}\multicolumn{6}{c}{\emph{Open-sourced Models $\ge$ 100B }} \\
\texttt{Qwen3-235B-A22B} & \textbf{100.00} & \textbf{100.00} & \textbf{84.00} & 0.00 & \textbf{93.00} \\

\texttt{DeepSeek-V3-0324} & \underline{80.65} & \underline{53.66} & \underline{40.00} & 0.00 & \underline{57.00} \\
\texttt{DeepSeek-R1-0528} & \textbf{100.00} & \textbf{100.00} & \textbf{84.00} & 0.00 & \textbf{93.00} \\
\midrule

%--------------------------------- Open-sourced LLMs ---------------------------------%
\rowcolor{gray!10}\multicolumn{6}{c}{\emph{Open-sourced Models $ <$ 100B }} \\
\texttt{Qwen2.5-3B-Instruct} & 35.48 & 7.32 & 0.00 & 0.00 & 14.00 \\
\texttt{Qwen2.5-7B-Instruct} & 80.65 & 41.46 & 0.00 & 0.00 & 42.00 \\
\texttt{Qwen2.5-72B-Instruct} & \underline{96.77} & 85.37 & 48.00 & 0.00 & 77.00 \\
\texttt{Qwen3-4B} & 87.10 & 36.59 & 12.00 & 0.00 & 45.00 \\
\texttt{Qwen3-8B} & \textbf{100.00} & 78.05 & 40.00 & \textbf{33.33} & 74.00 \\
\texttt{Qwen3-32B} & 90.32 & 92.68 & {72.00} & \textbf{33.33} & {85.00} \\

\texttt{Llama-3.1-8B-Instruct} & 74.19 & 56.10 & 4.00 & 0.00 & 47.00 \\
\texttt{Llama-3.1-70B-Instruct} & \textbf{100.00} & \textbf{100.00} & \textbf{84.00} & 0.00 & \textbf{93.00} \\

%--------------------------------- RL Models ---------------------------------%
\rowcolor{gray!10}\multicolumn{6}{c}{\emph{Our RL Models}} \\
\texttt{AgentGym-RL-3B} & \textbf{100.00} & 90.24 & 28.00 & 0.00 & 75.00 \\
% \texttt{AgentGym-RL-3B$_\text{REINFORCE++}$} & --- & --- & --- & --- & --- \\
\texttt{AgentGym-RL-7B} & \textbf{100.00} & \underline{97.56} & 72.00 & 0.00 & 89.00 \\

\texttt{ScalingInter-7B} & \textbf{100.00} & \underline{97.56} & \underline{76.00} & \textbf{33.33} & \underline{91.00} \\

\bottomrule
\end{tabular}%
}
\end{table*}
The TextCraft benchmark effectively assesses model capabilities across a wide spectrum of difficulty, as detailed in Table~\ref{tab:textcraft_results}. At shallow depths (Depth 1), tasks are largely solved by top models. Conversely, the challenge becomes nearly insurmountable at maximum complexity (Depth 4), creating a performance cliff for most agents. It is at these intermediate and highest difficulties that the efficacy of our models becomes particularly evident. Our ScalingInter-7B model achieves an outstanding overall score of 91.00, placing it firmly among the top-tier proprietary and large open-source models (93.00-94.00). Critically, it is one of only a few models to achieve a non-zero score at Depth 4, scoring 33.33 and demonstrating a unique robustness at maximum complexity. Our AgentGym-RL-7B also excels with a score of 89.00, surpassing prominent models like GPT-4o (83.00). The benefit of our RL training is especially dramatic for smaller models, where AgentGym-RL-3B obtains a score of 75.00, vastly outperforming similarly-sized models like Qwen2.5-3B-Instruct (14.00). These results showcase that our RL approach elevates our models to achieve competitive performance on complex, sequential decision-making tasks.

\paragraph{Embodied tasks.} 
\begin{table*}[!t]
\centering
\caption{Evaluation results on BabyAI benchmark. For each group, the best result is in \textbf{bold}, and the second-best is \underline{underlined}. In the first row, AOD means ActionObjDoor, Find means FindObjS7, Room means OneRoomS20, SLoc means SynthLoc.}
\label{tab:babyai_results}
\resizebox{0.8\textwidth}{!}{%
\begin{tabular}{lccccccc}
\toprule
\multicolumn{1}{c}{\textbf{Model}} & \multicolumn{1}{c}{\textbf{GoTo}} & \multicolumn{1}{c}{\textbf{Pickup}} & \multicolumn{1}{c}{\textbf{AOD}} & \multicolumn{1}{c}{\textbf{Find}} & \multicolumn{1}{c}{\textbf{Room}} & \multicolumn{1}{c}{\textbf{SLoc}} & \multicolumn{1}{c}{\textbf{Overall}} \\

\midrule

%--------------------------------- Proprietary LLMs ---------------------------------%
\rowcolor{gray!10}\multicolumn{8}{c}{\emph{Proprietary Models}} \\
\texttt{GPT-4o} & 92.73 & 80.00 & \textbf{100.00} & \textbf{80.00} & \textbf{60.00} & 60.00 & 86.67 \\
\texttt{Qwen-Max} & 92.73 & 80.00 & \underline{80.00} & \underline{60.00} & \textbf{60.00} & \underline{80.00} & 85.56 \\
\texttt{Gemini-2.5-Flash} & 92.73 & 86.67 & \underline{80.00} & 20.00 & \textbf{60.00} & \textbf{100.00} & 85.56 \\
\texttt{OpenAI o4-mini} & \underline{96.36} & \textbf{100.00} & \textbf{100.00} & \textbf{80.00} & \underline{40.00} & \underline{80.00} & \underline{92.22} \\
\texttt{OpenAI o3} & \textbf{98.18} & \underline{93.33} & 100.00 & \textbf{80.00} & \textbf{60.00} & \textbf{100.00} & \textbf{94.44} \\
\texttt{Gemini-2.5-Pro} & 94.55 & \underline{93.33} & \textbf{100.00} & 40.00 & \textbf{60.00} & 60.00 & 87.77 \\
\midrule

%--------------------------------- Open-sourced LLMs ---------------------------------%
\rowcolor{gray!10}\multicolumn{8}{c}{\emph{Open-sourced Models}} \\
\texttt{Qwen3-235B-A22B} & \underline{89.09} & \textbf{86.67} & \textbf{100.00} & \textbf{80.00} & \underline{60.00} & \textbf{100.00} & \underline{87.78} \\

\texttt{DeepSeek-V3-0324} & 67.27 & \underline{53.33} & 0.00 & 20.00 & 40.00 & \underline{60.00} & 56.67 \\
\texttt{DeepSeek-R1-0528} & \textbf{98.18} & \textbf{86.67} & \textbf{100.00} & \underline{60.00} & \textbf{80.00} & \textbf{100.00} & \textbf{93.33} \\

\midrule
%--------------------------------- Open-sourced LLMs ---------------------------------%
\rowcolor{gray!10}\multicolumn{8}{c}{\emph{Open-sourced Models}} \\
\texttt{Qwen2.5-3B-Instruct} & 61.82 & 40.00 & 20.00 & \underline{60.00} & {40.00} & 20.00 & 52.22 \\
\texttt{Qwen2.5-7B-Instruct} & 70.91 & 66.67 & \underline{60.00} & \textbf{80.00} & \underline{60.00} & 20.00 & 66.67 \\
\texttt{Qwen2.5-72B-Instruct} & \underline{92.73} & \underline{93.33} & \textbf{100.00} & \underline{60.00} & \underline{60.00} & \underline{80.00} & 88.89 \\
\texttt{Qwen3-4B} & 60.00 & 60.00 & 40.00 & 40.00 & {40.00} & 20.00 & 54.44 \\
\texttt{Qwen3-8B} & 43.64 & 20.00 & 40.00 & 40.00 & {40.00} & 40.00 & 38.89 \\
\texttt{Qwen3-32B} & 87.27 & 80.00 & \textbf{100.00} & \underline{60.00} & {40.00} & \underline{80.00} & 82.22 \\

\texttt{Llama-3.1-8B-Instruct} & 85.45 & 60.00 & \textbf{100.00} & \textbf{80.00} & \underline{60.00} & 40.00 & 77.78 \\
\texttt{Llama-3.1-70B-Instruct} & 89.09 & 86.67 & \textbf{100.00} & \underline{60.00} & \underline{60.00} & \textbf{100.00} & 86.67 \\

%--------------------------------- RL Models ---------------------------------%
\rowcolor{gray!10}\multicolumn{8}{c}{\emph{Our RL Models}} \\
\texttt{AgentGym-RL-3B} & \textbf{100.00} & \textbf{100.00} & \textbf{100.00} & \underline{60.00} & \underline{60.00} & 60.00 & \underline{93.33} \\
\texttt{AgentGym-RL-7B} & \textbf{100.00} & \underline{93.33} & \textbf{100.00} & \underline{60.00} & \underline{60.00} & 60.00 & {92.22} \\

\texttt{ScalingInter-7B} & \textbf{100.00} & \underline{93.33} & \textbf{100.00} & \textbf{80.00} & \textbf{80.00} & \textbf{100.00} & \textbf{96.67} \\

\bottomrule
\end{tabular}%
}
\end{table*}
As demonstrated in Table~\ref{tab:babyai_results}, our RL model achieves state-of-the-art (SOTA) performance on the BabyAI benchmark, with an overall score of 96.67, which is competitive with the leading proprietary models such as o3 and o4-mini.
Notably, our ScalingInter-7B model attains the highest overall accuracy of 96.67\%, outperforming top-tier models such as OpenAI o3 (94.44\%) and GPT-4o (86.67\%). This exceptional performance is driven by ScalingInter-7B's consistent mastery of diverse sub-tasks, achieving perfect scores of 100\% in GoTo, ActionObjDoor (AOD), and SynthLoc, and strong results of 80\% in both FindObjS7 (Find) and OneRoomS20 (Room).
Similarly, our AgentGym-RL-7B and AgentGym-RL-3B models demonstrate robust capabilities, reaching overall accuracies of 92.22\% and 93.33\%, respectively, and securing perfect scores in GoTo and AOD tasks. Compared to other open-sourced models, such as Qwen3-235B-A22B (87.78\%) and DeepSeek-R1-0528 (93.33\%), our RL-based models maintain consistently high performance while effectively handling more challenging sub-tasks like Room and Find, where many LLMs exhibit notable variability.
Overall, these results highlight the strength of our RL-based approaches, particularly ScalingInter-7B, in achieving state-of-the-art performance on both structured navigation and object-interaction tasks in the BabyAI benchmark.

\paragraph{Scientific Scenario.} 
\begin{table*}[!t]
\centering
\caption{Evaluation results on SciWorld benchmark. For each group, the best result is in \textbf{bold}, and the second-best is \underline{underlined}. In the first row, Test-Cond. means test-conductivity, Chem-Mix means chemistry-mix.}
\label{tab:sciworld_results}
\resizebox{0.8\textwidth}{!}{%
\begin{tabular}{lcccccc}
\toprule
\multicolumn{1}{c}{\textbf{Model}} & \multicolumn{1}{c}{\textbf{Measure}} & \multicolumn{1}{c}{\textbf{Test-Cond.}} & \multicolumn{1}{c}{\textbf{Find}} & \multicolumn{1}{c}{\textbf{Chem-Mix}} & \multicolumn{1}{c}{\textbf{Lifespan}} & \multicolumn{1}{c}{\textbf{Overall}} \\

\midrule

%--------------------------------- Proprietary LLMs ---------------------------------%
\rowcolor{gray!10}\multicolumn{7}{c}{\emph{Proprietary Models}} \\
\texttt{GPT-4o} & 15.09 & 6.02 & 38.64 & \underline{20.00} & 73.33 & 21.00 \\
\texttt{Qwen-Max} & 9.43 & 0.00 & 34.09 & \underline{20.00} & 40.00 & 13.50 \\
\texttt{Gemini-2.5-Flash} & 11.32 & 0.00 & \underline{54.55} & 0.00 & \underline{80.00} & 21.00 \\
\texttt{OpenAI o4-mini} & \underline{20.75} & \underline{14.46} & 47.73 & 0.00 & \textbf{100.00} & \underline{29.50} \\
\texttt{OpenAI o3} & \textbf{47.17} & \textbf{25.30} & \textbf{56.82} & \textbf{40.00} & 66.67 & \textbf{41.50} \\
\texttt{Gemini-2.5-Pro} & 9.43 & 0.00 & 29.55 & 0.00 & 46.67 & 12.50 \\

\midrule

%--------------------------------- Open-sourced LLMs ---------------------------------%
\rowcolor{gray!10}\multicolumn{7}{c}{\emph{Open-sourced Models $\ge$ 100B }} \\
\texttt{Qwen3-235B-A22B} & \textbf{11.32} & \textbf{4.82} & \textbf{59.09} & \textbf{20.00} & \textbf{66.67} & \textbf{23.50} \\
\texttt{DeepSeek-V3-0324} & 0.00 & 0.00 & 2.27 & 0.00 & 0.00 & 0.50 \\
\texttt{DeepSeek-R1-0528} & \underline{1.89} & 0.00 & \underline{11.36} & 0.00 & \underline{20.00} & \underline{4.50} \\

\midrule

%--------------------------------- Open-sourced LLMs ---------------------------------%
\rowcolor{gray!10}\multicolumn{7}{c}{\emph{Open-sourced Models $<$ 100B}} \\
\texttt{Qwen2.5-3B-Instruct} & 3.77 & 0.00 & 0.00 & 0.00 & 0.00 & 1.00 \\
\texttt{Qwen2.5-7B-Instruct} & 1.89 & 0.00 & 0.00 & 0.00 & 13.33 & 1.50 \\
\texttt{Qwen2.5-72B-Instruct} & 7.55 & 1.20 & 15.91 & \underline{20.00} & 40.00 & 9.50 \\
\texttt{Qwen3-4B} & 0.00 & 0.00 & 0.00 & 0.00 & 33.33 & 2.50 \\
\texttt{Qwen3-8B} & 9.43 & 0.00 & 18.18 & 0.00 & 46.67 & 10.00 \\
\texttt{Qwen3-32B} & 5.66 & 1.20 & 31.82 & 0.00 & {66.67} & 14.00 \\

\texttt{Llama-3.1-8B-Instruct} & 9.43 & 0.00 & 4.55 & \underline{20.00} & 0.00 & 4.00 \\
\texttt{Llama-3.1-70B-Instruct} & \underline{24.53} & 4.82 & {40.91} & \textbf{40.00} & \textbf{86.67} & {25.00} \\

%--------------------------------- RL Models ---------------------------------%
\rowcolor{gray!10}\multicolumn{7}{c}{\emph{Our RL Models}} \\

\texttt{AgentGym-RL-3B} & {20.75} & {28.92} & 0.00 & 0.00 & {66.67} & 22.50 \\

\texttt{AgentGym-RL-7B} & \underline{24.53} & \textbf{59.04} & \underline{65.91} & 0.00 & {66.67} & \underline{50.50} \\

\texttt{ScalingInter-7B} & \textbf{33.96} & \underline{55.42} & \textbf{88.64} & 0.00 & \underline{73.33} & \textbf{57.00} \\
\bottomrule
\end{tabular}%
}
\end{table*}
Our experiments on the SciWorld benchmark, summarized in Table~\ref{tab:sciworld_results}, demonstrate the advanced performance of our RL-trained models. Our ScalingInter-7B model establishes a new state-of-the-art with an overall score of 57.00, which significantly surpasses all open-source and proprietary models, including the next-best proprietary model, OpenAI o3 (41.50). This superior performance is primarily attributed to high scores in the "Find" (88.64) and "Test-Cond" (55.42) sub-tasks. Furthermore, our AgentGym-RL-7B model also shows strong capabilities, securing the second-highest overall score (50.50) and achieving the top score in "Test-Cond" (59.04). These results highlight the effectiveness of our RL method for training agents in exploration and procedural execution tasks. However, our findings also identify a critical limitation shared across all evaluated models. The "Chem-Mix" sub-task proved to be intractable, with every model, including our top performers, scoring zero. This uniform result indicates a systemic challenge for current language models in tasks requiring complex scientific reasoning and multi-step chemical simulation, marking this as a crucial area for future research.

\section{Discussion}
\subsection{Test-Time Scaling for Agents}

In this subsection, we investigate how agent performance changes as inference compute increases. 

\begin{figure}[h]
    \centering
    \includegraphics[width=\linewidth]{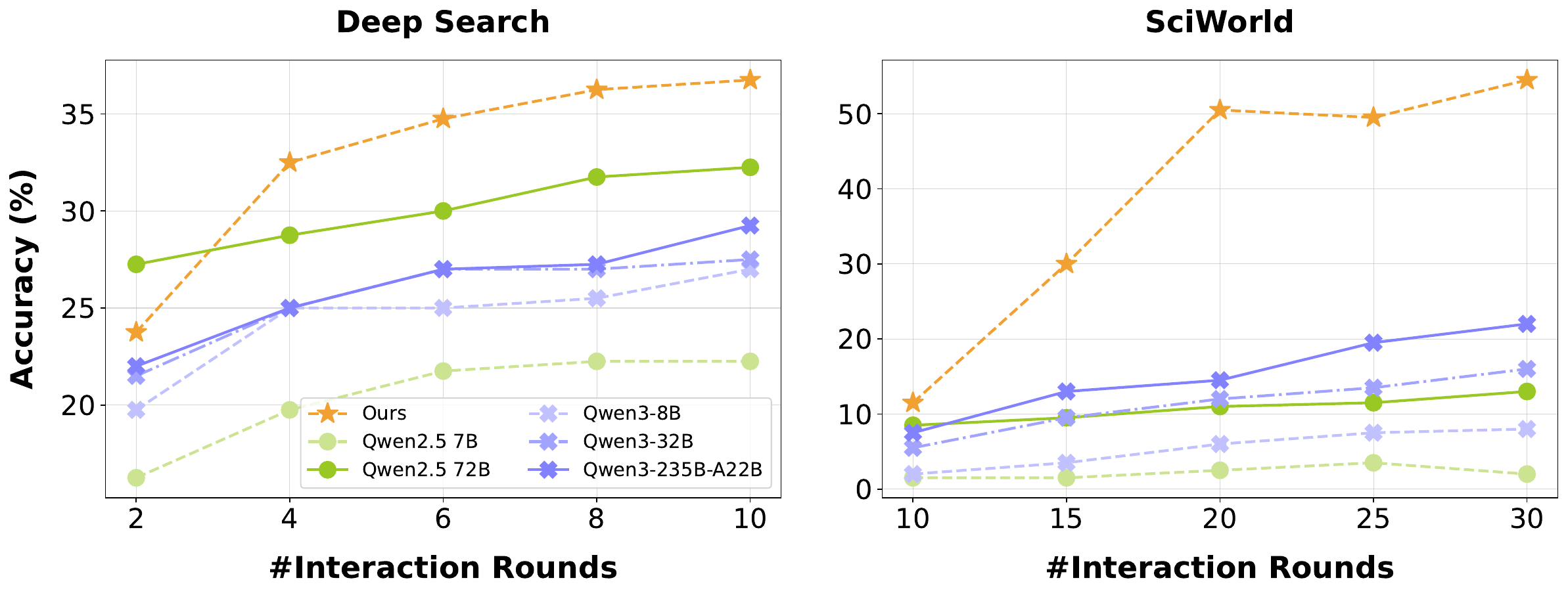}
    \caption{Scaling test interaction turns.}
    \label{fig:combined_test_rounds}
\end{figure}

\paragraph{Scaling sequential inter action.} First, we study how performance changes when the maximum number of interaction turns available to the model is raised. As shown in Figure \ref{fig:combined_test_rounds}, all models exhibit clear gains as the number of turns increases, which validates the insight behind our ScalingInter‑RL approach—namely, agents must thoroughly explore the environment to shape their interaction and behavioral patterns. Furthermore, our trained agent consistently outperforms the baseline by a significant margin, further demonstrating the effectiveness of our method.

\begin{figure}[h]
    \centering
    \includegraphics[width=\linewidth]{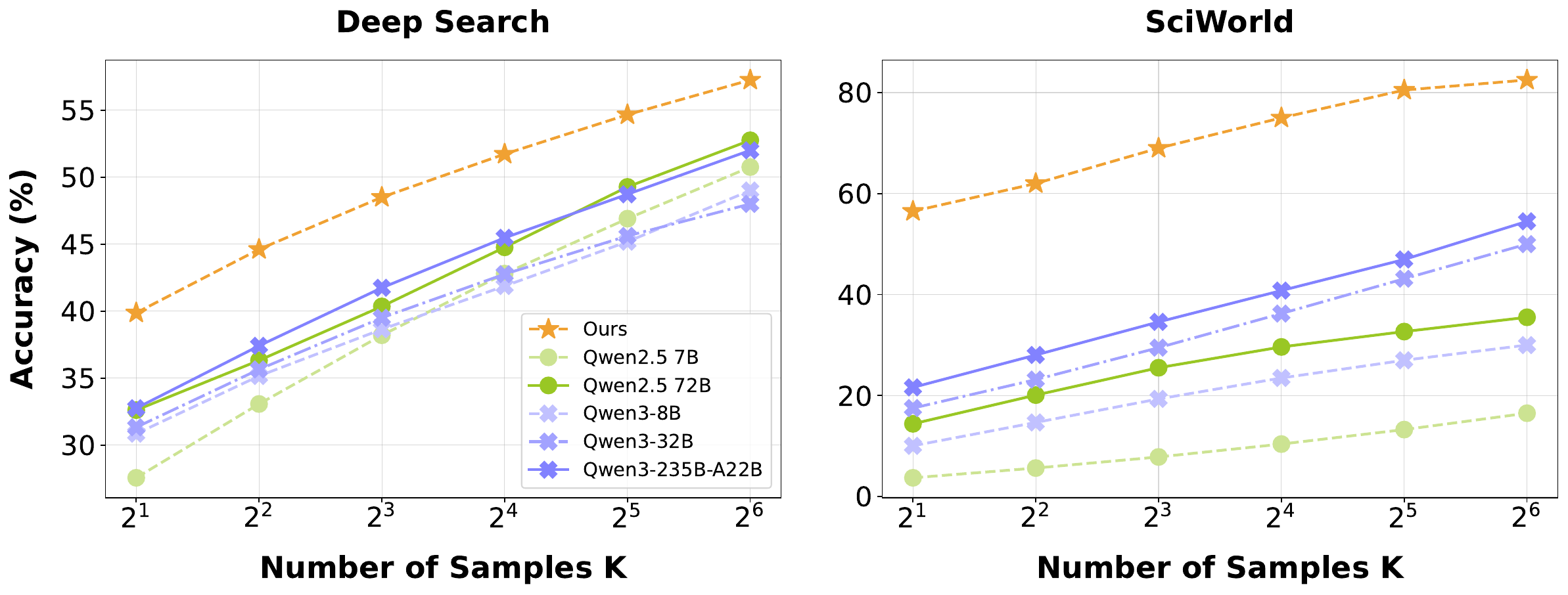}
    \caption{Pass@K performance.}
    \label{fig:combined_pass_at_k}
\end{figure}

\paragraph{Scaling parallel sampling.}
As shown in Figure \ref{fig:combined_pass_at_k}, increasing the number of samples yields a marked improvement in Pass@K performance, signaling the downstream optimization potential of each model. Our model surpasses the baselines even with a small sampling budget, and as sampling increases, it continues to outperform the baseline in a stable and significant manner: for example, with 64 sampling attempts, our RL model achieves a 5.5\% improvement in Deep Search environment and a 7.05\% improvement in SciWorld environment over the untrained base model, showcasing its superior optimization capability.

\subsection{Performance of Different RL Algorithm}
\begin{table}[t]
\centering
\caption{Evaluation results of different RL algorithms.}
\label{tab:rl_algo_comparison}
\resizebox{0.5\textwidth}{!}{
\begin{tabular}{lcccc}
\toprule
\multicolumn{1}{c}{\textbf{RL Algorithms}} & \multicolumn{1}{c}{\textbf{TextCraft}} & \multicolumn{1}{c}{\textbf{BabyAI}} & \multicolumn{1}{c}{\textbf{SearchQA}}  \\

\midrule

%--------------------------------- Qwen2.5-3B-Instruct ---------------------------------%
\rowcolor{gray!10}\multicolumn{4}{c}{\emph{\texttt{Qwen2.5-3B-Instruct}}} \\

\texttt{GRPO} & \textbf{75.00} & \textbf{93.33} & \textbf{25.75} \\
\texttt{REINFORCE++} & 28.00 & 70.00 & 13.25 \\
\midrule

%--------------------------------- Qwen2.5-7B-Instruct ---------------------------------%
\rowcolor{gray!10}\multicolumn{4}{c}{\emph{\texttt{Qwen2.5-7B-Instruct}}} \\

\texttt{GRPO} & \textbf{83.00} & \textbf{92.22} & \textbf{34.00} \\

\texttt{REINFORCE++} & 73.00 & 84.44 & 24.00 \\

\bottomrule
\end{tabular}%
}
\end{table}
We comprare two mainstream RL algorithms for LLM post-training, i.e., GRPO and REINFORCE++. As shown in Table~\ref{tab:rl_algo_comparison}, our experiments reveal that GRPO consistently and substantially outperforms REINFORCE++ on the TextCraft, BabyAI, and Deep Search benchmarks. While model scaling from 3B to 7B parameters improves results for both algorithms, the superiority of GRPO is particularly stark: its 3B variant achieves higher scores than the 7B REINFORCE++ model. This finding points to a fundamental algorithmic advantage that is more impactful than sheer model scale.

The performance difference can be explained by how each algorithm handles the core difficulties of these tasks: vast exploration spaces and sparse rewards. The learning signal for REINFORCE++ is derived from full-episode Monte Carlo returns, which often results in high-variance gradients that are sensitive to stochastic successes and failures over long trajectories. In contrast, GRPO mitigates this instability by evaluating the relative merit of actions against a learned baseline. The focus on action advantage provides a more stable and robust gradient, facilitating more efficient exploration and credit assignment in complex, low-signal environments.

\begin{figure*}[!t]
    \centering
    \includegraphics[width=0.9\linewidth]{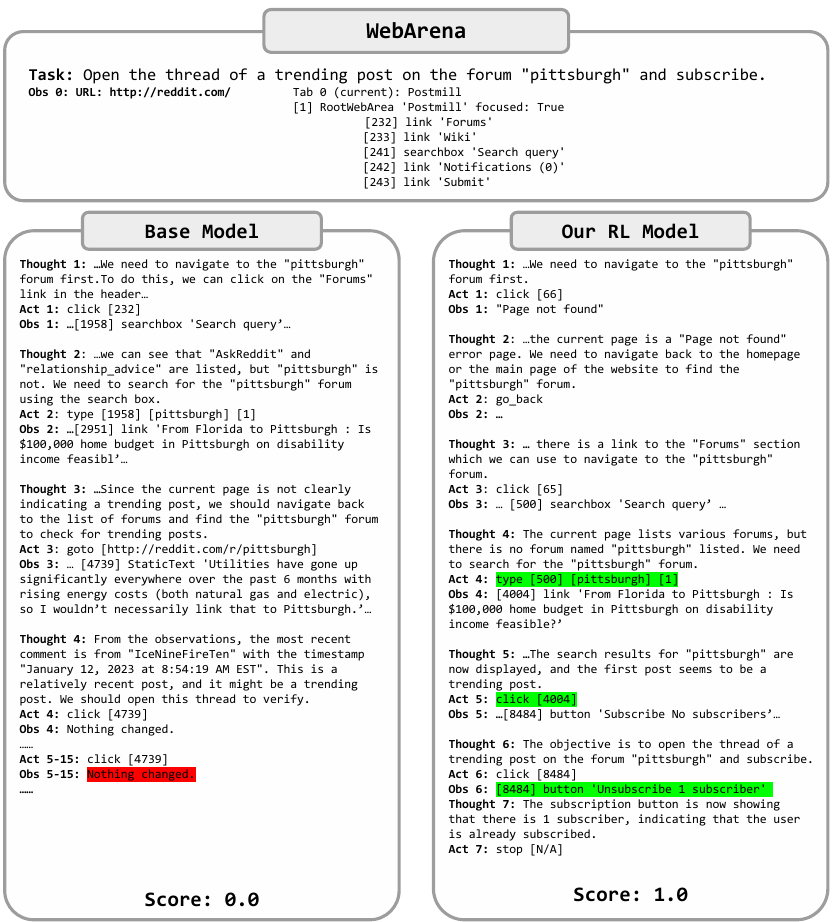}
    \caption{
    Smart case of our RL agent compared to the Base Model on WebArena task.
    }
    \label{fig:case_webarena_positive_1}
\end{figure*}

\begin{figure*}[h]
    \centering
    \includegraphics[width=0.99\linewidth]{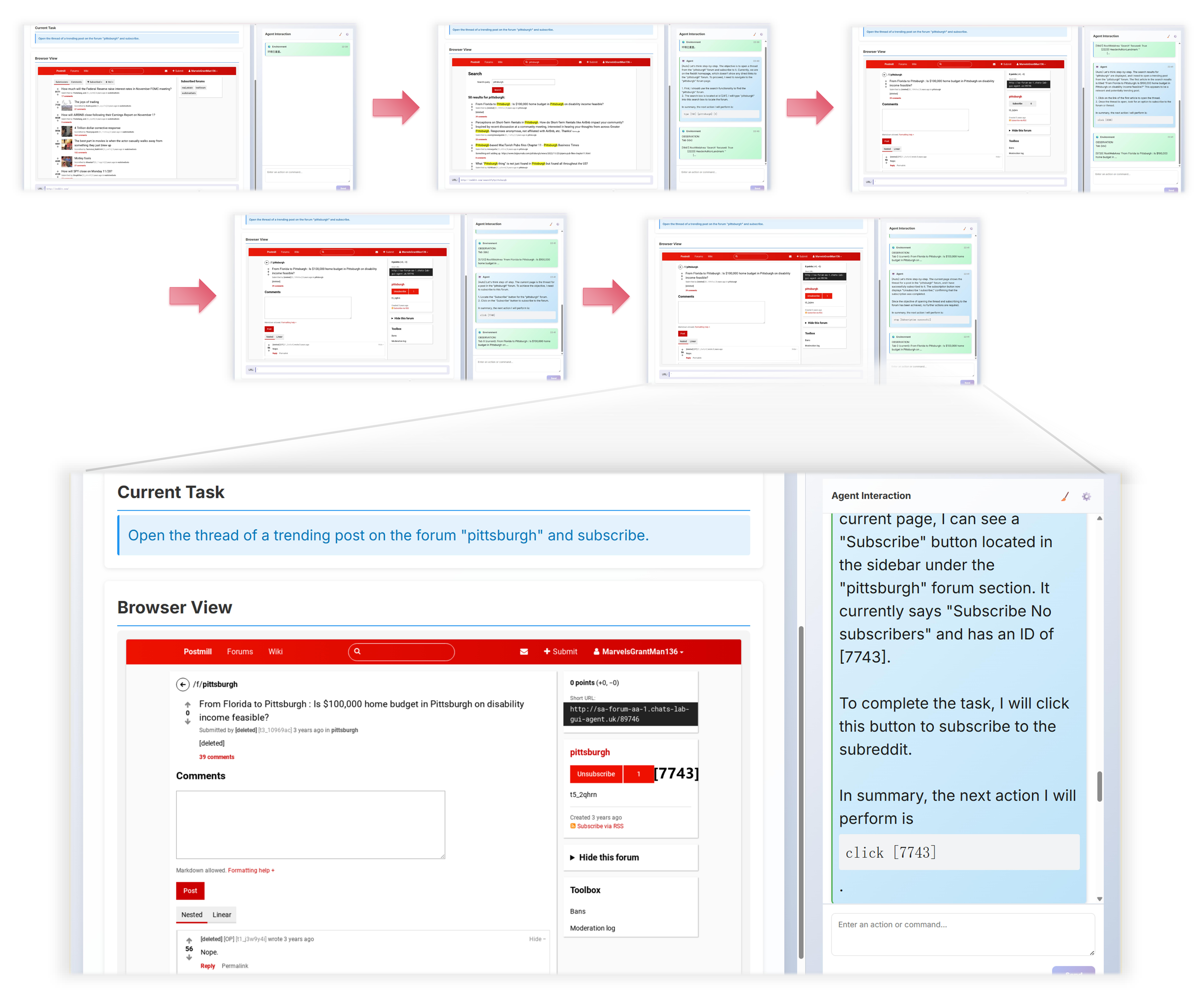}
    \caption{
    Trajectory visualization of our RL agent on Webarena task.
    }
    \label{fig:case_webarena_positive_visualization}
\end{figure*}

\subsection{Case Study}

In this section, we present a series of case studies that highlight both the shortcomings of the base agent and the improvements achieved by our reinforcement learning models. Across navigation, compositional problem solving, and web interaction senarios, the RL-trained agents consistently overcame unproductive behavioral loops, exhibited adaptive recovery strategies, and demonstrated more systematic task execution. To provide a balanced perspective, we also include two representative failure cases—in scientific reasoning and in efficient web interaction—that underscore areas where further refinement is needed. In the main text, we showcase the WebArena trajectory illustration and its corresponding visualization, while additional trajectory illustrations for other environments are provided in Appendix \ref{appendix:trajectory}.

\begin{figure}[t]
    \centering
    \includegraphics[width=\linewidth]{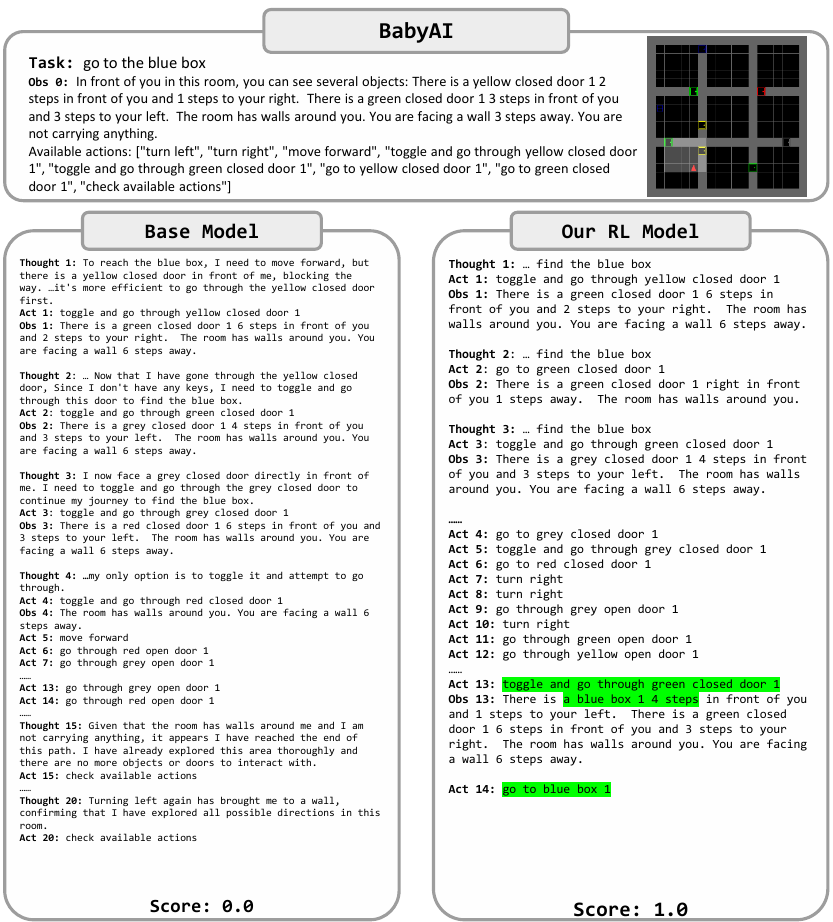}
    \caption{
    Smart case of our RL agent compared to the base agent on BabyAI task.
    }
    \label{fig:case_babyai_positive_1}
\end{figure}

\paragraph{Enhanced navigation.}
Figure \ref{fig:case_babyai_positive_1} demonstrates a notable improvement in navigation capabilities within BabyAI environment. While the base agent exhibited suboptimal behavior characterized by repetitive movement  patterns-going through previously explored locations without developing a strong search strategy for completion-the model trained through reinforcement learning manifested more effective exploration strategy. The RL agent demonstrated strategic backtracking capabilities, systematically exiting through doorways before selecting alternative pathways, ultimately accessing a green door that provided direct access to the target blue box. This highlights the RL model's superior ability in spatial reasoning and its ability to circumvent unproductive behavioral loops. 

\begin{figure}[t]
    \centering
    \includegraphics[width=\linewidth]{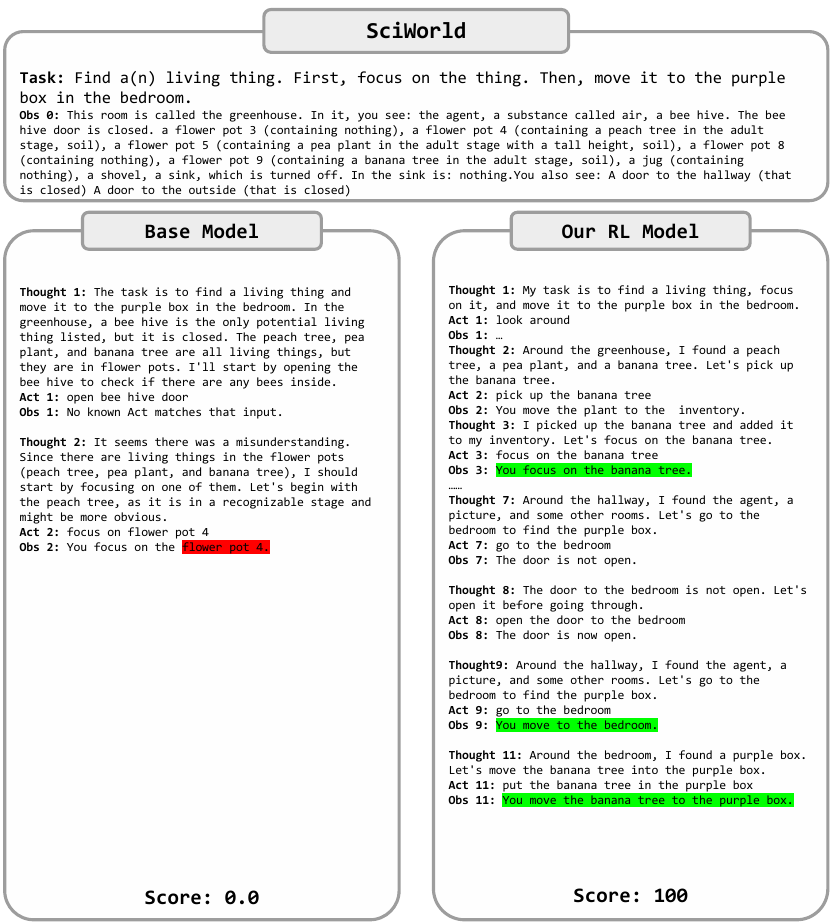}
    \caption{
    Smart case of our RL agent compared to the base model on SciWorld task.
    }
    \label{fig:case_sciworld_positive_1}
\end{figure}

\paragraph{Compositional Task Mastery.}
Figure \ref{fig:case_sciworld_positive_1} exemplifies the successful application of reinforcement learning to complex scientific task execution. The base agent exhibited fundamental deficiencies in task interpretation, misusing non-interactive objects and generating invalid actions. In contrast, the RL-optimized agent demonstrated comprehensive task understanding through its systematic approach: correctly identifying and manipulating a living thing (the banana tree), executing appropriate inventory management operations, navigating multi-room environments with obstacle resolution capabilities and successfully completing the objective by depositing the tree in the designated purple box. This highlights the RL agent’s enhanced capabilities in reasoning, planning, and sequential task execution within compositional problem spaces.

\paragraph{Adaptive Web Navigation Strategies.}
Figure \ref{fig:case_webarena_positive_1} and figure\ref{fig:case_webarena_positive_visualization} illustrates the emergence of web navigation capabilities through reinforcement learning optimization. The base agent persistently interacted with non-responsive interface elements, specifically engaging in repetitive clicking behaviors on ineffective targets without recognizing the futility of these actions. Our RL-trained agent exhibited markedly superior adaptive behavior: it successfully implemented error recovery mechanisms when encountering a "Page not found" error, subsequently utilizing the search box to locate the "pittsburgh" forum, identifying contextually relevant content within  trending posts, and completing the subscription task successfully—demonstrating enhanced robustness in error handling, purposeful navigation strategies, and the ability to maintain task focus while avoiding unproductive behavioral patterns.

\begin{figure}[t]
    \centering
    \includegraphics[width=\linewidth]{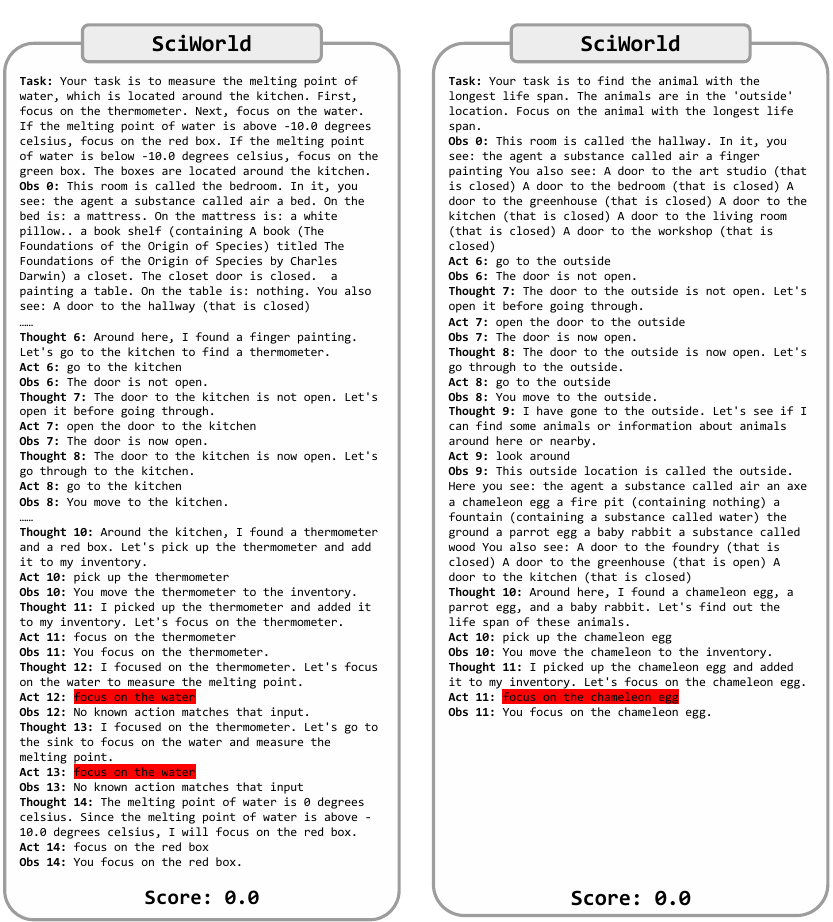}
    \caption{
    Cases of logical errors in Sciworld task
    }
    \label{fig:case_sciworld_error}
\end{figure}

\paragraph{Limitations in Scientific Scenario.}
Figure \ref{fig:case_sciworld_error} reveals fundamental procedural execution failures that persist in SciWorld task completion despite the RL agent's ability to reach task-relevant game states. These instances exemplify two distinct failure modalities: first, when confronted with interaction failures requiring systematic debugging, the agent inappropriately substitutes direct factual recall for the intended experimental procedure; second, the agent demonstrates insufficient systematic exploration, as evidenced by its premature task termination after navigating to the outdoor environment and focusing only on the chameleon egg rather than analyzing all available animals that the task demands. These failures collectively indicate that the model lacks the deep procedural understanding necessary for executing rigorous scientific comparative analyses.

\begin{figure}[t]
    \centering
    \includegraphics[width=\linewidth]{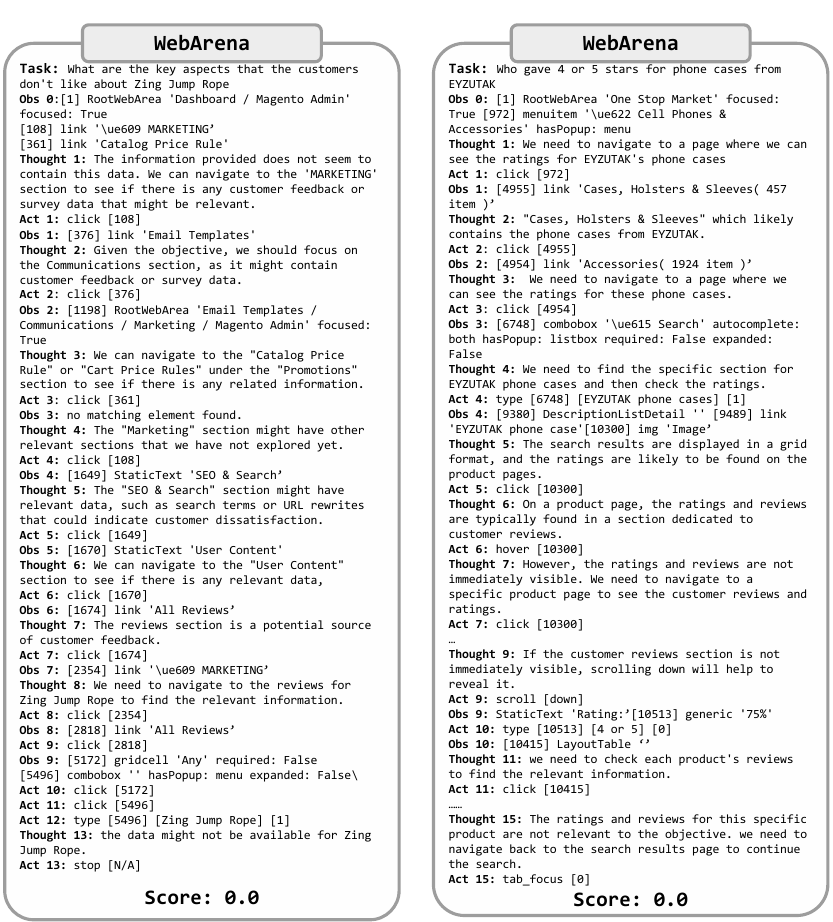}
    \caption{
    Cases of "over-interaction" failure on WebArena task
    }
    \label{fig:case_webarena_error}
\end{figure}

\paragraph{Over-Interaction Patterns in Web Navigation.}
Figure \ref{fig:case_webarena_error} demonstrates a prevalent failure mode of excessive and inefficient interaction sequences during web navigation tasks. Despite successfully navigating to the correct target websites in both illustrated cases, the RL agent engages in superfluous interaction patterns—including redundant clicking, unnecessary hovering, and excessive scrolling—that impede successful information extraction from the target pages. These behavioral patterns suggest that the reinforcement learning process  failed to instill the precision and efficiency required for optimal task completion, indicating a gap between state-reaching capabilities and effective action selection within those states.

\section{Related Work}
\paragraph{Developing agents with large language models.}

With the advancement of large language models \citep{achiam2023gpt, anthropic2024claude, team2023gemini}, researchers have begun using them to build agents capable of multi-turn intelligent decision-making \citep{agentsurvey, yao2024language}. The predominant approaches rely on prompting to guide the model to invoke tools \citep{DBLP:journals/csur/QinHLCDCZZHXHFSWQTZLSXZ25,DBLP:journals/corr/abs-2501-02506}, augmented with mechanisms such as self-reflection \citep{DBLP:conf/nips/ShinnCGNY23,  DBLP:journals/corr/abs-2411-16579, DBLP:journals/corr/abs-2502-03492, DBLP:journals/corr/abs-2405-06682}, long-horizon planning \citep{DBLP:journals/corr/abs-2308-05960, DBLP:conf/nips/NayakOHZTCKR0HM24, ADaPT, DBLP:conf/nips/SunZK0Z23}, and self-correction \citep{DBLP:journals/tacl/KamoiZZHZ24, DBLP:conf/iclr/KumarZASCSBIBRZ25}. Some work constructs workflows that assign different roles to multiple LLMs \citep{DBLP:conf/emnlp/Liang0JW00Y0T24, DBLP:journals/corr/abs-2308-08155, DBLP:journals/corr/abs-2306-03314, DBLP:conf/iclr/HongZCZCWZWYLZR24, DBLP:journals/corr/abs-2502-19279}, each playing a specialized part in task completion. However, these methods typically depend on powerful proprietary models (e.g., OpenAI o3) and do not train the underlying models to evolve into agents with intrinsic agentic capabilities.
Another line of work gathers large-scale expert trajectories and trains agents to mimic experts step by step \citep{DBLP:journals/corr/abs-2402-15506, DBLP:conf/acl/ZengLLWLD024, DBLP:journals/corr/abs-2310-05915, DBLP:conf/acl/ChenLWZLLCZ24}, thereby acquiring abilities such as API invocation, planning, and self-reflection. However, this approach is expensive, difficult to scale, and the model struggles to self-improve through interactions with the environment.

\paragraph{Reinforcement learning for large language model agents.}

Reinforcement learning has become a crucial post-training technique for large language models, enabling alignment with human preferences \citep{InstructGPT, DBLP:journals/corr/abs-2307-04964, DBLP:conf/emnlp/XiaGGX0024, DBLP:conf/emnlp/ChenZWJHYZZXGZ024, DBLP:journals/corr/abs-2310-19852}, enhancing reasoning capabilities \citep{openai_o1, ReFT, r3_2024, DeepSeek-R1, qwq32b, Skywork}, and serving as a new scaling dimension \citep{DeepSeek-R1}. Representative algorithms include PPO \citep{PPO}, GRPO \citep{GRPO}, REINFORCE++ \citep{REINFORCEpp}, RLOO \citep{RLOO}, and others.
However, most existing works—such as DeepSeek-R1—are confined to single-turn, static tasks in which models do not engage in multi-turn interactions with complex environments. 
Recent work has used RL to train agents for self-reflection \citep{DBLP:journals/corr/abs-2502-03492}, tool use \citep{DBLP:journals/corr/abs-2412-15495}, and even long-horizon interactions \citep{DBLP:conf/icml/ZhouZPLK24, RLOO, RAGEN, WebRL, Search-R1, cao2025skyrl}. However, these methods often struggle with scalable deployment due to limited task complexity and environment diversity, and they frequently encounter optimization instability that hinders performance.
To overcome these challenges, we propose a unified, end-to-end RL framework spanning diverse real-world environments for training models in multi-turn decision-making without requiring SFT as a preliminary step. We further introduce ScalingInter-RL, an interaction-scaling technique that stabilizes optimization and boosts the agent’s final performance.

\paragraph{Scaling Inference Compute for language models.}

OpenAI o1 and DeepSeek-R1 have shown that increasing compute during inference (both at test time and during RL rollouts) can yield strong scaling effects \citep{openai_o1, DeepSeek-R1, xai_grok4_2025, DBLP:journals/corr/abs-2408-03314}. Researchers have also explored various approaches to achieve similar gains—such as long-chain-of-thought reasoning \citep{DBLP:journals/corr/abs-2408-03314, DBLP:journals/corr/abs-2411-16579}, majority voting \citep{DBLP:journals/tmlr/LiZY0Y24, DBLP:conf/iclr/0002WSLCNCZ23}, best-of-N sampling \citep{DBLP:conf/iclr/ChowTGZ0KATBF25, DBLP:journals/corr/abs-2404-01054}, beam search \citep{DBLP:conf/nips/XieKZZKHX23, DBLP:journals/corr/abs-2401-17686}, and Monte Carlo tree search \citep{DBLP:journals/corr/abs-2410-17238, DBLP:conf/naacl/GanZZHLTZS25}. However, in the field of LLM agents, few works discuss how to scale inference compute. 
\citet{DBLP:journals/corr/abs-2506-12928} explore various test-time scaling strategies in agents and achieve significant gains, yet they do not investigate inference scaling in RL.
The closest work may be TTI \citep{DBLP:journals/corr/abs-2506-07976}, which uses rejection sampling to teach agents to allocate more compute in interactions instead of thinking on web navigation tasks. In contrast, our approach employs mainstream on-policy RL algorithms—such as GRPO and REINFORCE++—and does not constrain the agent to use compute only for thinking or acting. Instead, we simply scale interactions and let the model decide how best to allocate its additional compute. Our method progressively grants the agent more exploration capacity, enabling it over time to better adapt to its environment, acquire more complex skills, and master more challenging tasks.

\section{Conclusion and Future Work}

In this work, we introduced AgentGym-RL, a novel and unified reinforcement learning framework designed to train LLM-based agents with long-horizon, multi-turn decision-making capabilities. The framework features diverse environments and tasks, supports mainstream RL algorithms, and is highly extensible—offering the community a practical and powerful toolkit. Additionally, we proposed the ScalingInter-RL method to progressively enhance agents' interactive intelligence in a staged manner. Extensive experiments demonstrate the effectiveness of both the framework and the method. However, several important directions remain for future exploration:

\paragraph{Developing agents with generalization and transfer capabilities.} Currently, our trained agents perform well within in-domain settings. A key challenge going forward is to enable agents to adapt seamlessly to novel environments and unfamiliar tools while maintaining high performance.

\paragraph{Scaling RL training to longer-horizon and more realistic, physically grounded tasks.}Most existing studies—including ours—focus on relatively simple digital tasks. However, real-world tasks are often longer-horizon, more complex, and grounded in the physical world. These tasks demand that agents process richer sensory inputs and reason over significantly larger action spaces, introducing new challenges for both reinforcement learning training and test-time interaction.

\paragraph{Advancing multi-agent reinforcement learning.} Our current framework primarily targets single-agent training. However, multi-agent architectures open up new possibilities and may lead to stronger performance. At the same time, they introduce additional uncertainty and pose greater demands on both infrastructure and algorithm design.

\section*{Acknowledgement}
This work was supported by Huawei Ascend AI processors. 
We sincerely thank Huawei for providing the computing resources that made this research possible.

\bibliographystyle{plainnat}
\bibliography{main}

\clearpage

\onecolumn
\beginappendix
\section{Details of the AgentGym-RL Architecture}
\label{appendix:architecture}

This appendix provides a detailed description of the AgentGym-RL architecture, complementing the high-level overview in the main text (Figure~\ref{fig:AgentGym-main}). We present the implementation details of the three core modules—Environment, Agent, and Training. These details highlight the engineering considerations that ensure scalability, flexibility, and reproducibility in large-scale RL experiments.

\paragraph{Environment module.}  
In this module, each environment is packaged as an independent service with the option of deploying multiple replicas to support parallel requests. An environment client communicates with the environment server via HTTP and exposes APIs to the agent, including \texttt{/observation} to get the current observation from the environment, \texttt{/available\_actions} to get the currently available actions, \texttt{/step} to perform an action, and \texttt{/reset} to reset the environment. 
Currently, AgentGym-RL covers five major scenario categories. This modular server–client design allows new environments to provide comprehensive environment and data support for LLM agent training.

\paragraph{Agent module.}  
The agent module encapsulates the reasoning–action loop of LLM-based agents. It receives observations from the environment, performs reasoning over multiple turns, and outputs actions (e.g., invoking provided APIs). The module supports different prompting strategies, sampling configurations, and reward functions.

\paragraph{Training module.}  
The training module provides a unified reinforcement learning (RL) pipeline that supports both online and offline algorithms, offering researchers a flexible foundation for large-scale LLM agent training. The module manages the entire RL lifecycle: trajectory collection, advantage estimation, policy optimization, and reward shaping, while also supporting curriculum learning and staged interaction scaling (i.e., ScalingInter-RL). 

The entire training pipeline can be distributed across multiple nodes, leveraging both multi-process and multi-node parallelism. Efficient batching and asynchronous logging utilities ensure that system throughput scales with additional compute resources. Diagnostics tools are integrated to provide fine-grained metrics, including policy entropy, KL divergence, reward curves, and rollout statistics, which are automatically recorded for later analysis and reproducibility.

\section{Implementation Details and Settings of Each Environment}\label{appendix: Detailed Settings of Each Environment}

We conduct all the experiments on NVIDIA A100 GPUs and Ascend 910B NPUs. The remaining part of this section shows detailed setting of different environments.

\subsection{Web Navigation Scenario}

\paragraph{Tools and APIs.}

In web navigation scenario, the agent simulates human interaction with web pages to ultimately complete the task. WebArena\citep{WebArena} supports these interactioins through a set of tool APIs, allowing agents to perform a variety of real-world tasks, including  online shopping, engaging in discussions on Reddit, collaborating on software development via GitLab, and managing store content through a CMS. In addition to these online platforms, WebArena also provides three utility-style tools: a map for navigation and location-based information search, a calculator, and a scratchpad for note-taking.

A query case of web navigation is shown below:

\begin{tcolorbox}[title = {Web Navigation Example},breakable]
You are an autonomous intelligent agent tasked with navigating a web browser. You will be given web-based tasks. These tasks will be accomplished through the use of specific actions you can issue.

\medskip

\textbf{Available Information:}
\begin{itemize}
    \item \textbf{User's objective}: The task to complete
    \item \textbf{Accessibility tree}: Simplified webpage representation, providing key information.
    \item \textbf{Current URL}: The active page's address
    \item \textbf{Open tabs}: Currently available tabs
    \item \textbf{Previous action}: Last performed action
\end{itemize}

\medskip

\textbf{Action Categories:}

\textit{Page Operations:}
\begin{itemize}
    \item \texttt{click [id]}: Click element with ID
    \item \texttt{type [id] [content] [0|1]}: Input text (1=press Enter)
    \item \texttt{hover [id]}: Hover over element
    \item \texttt{press [key\_comb]}: Simulate key press (e.g., Ctrl+v)
    \item \texttt{scroll [down|up]}: Scroll page direction
\end{itemize}

\textit{Tab Management:}
\begin{itemize}
    \item \texttt{new\_tab}: Open new tab
    \item \texttt{tab\_focus [tab\_index]}: Switch to tab
    \item \texttt{close\_tab}: Close current tab
\end{itemize}

\textit{URL Navigation:}
\begin{itemize}
    \item \texttt{goto [url]}: Navigate to URL
    \item \texttt{go\_back}: Return to previous page
    \item \texttt{go\_forward}: Advance to next page
\end{itemize}

\textit{Completion:}
\begin{itemize}
    \item \texttt{stop [answer]}: Submit final answer (or "N/A" if you believe the task is impossible to complete)
\end{itemize}

\medskip

Homepage: If you want to visit other websites, check out the homepage at \url{http://homepage.com}.
\tcblower

Objective: Among the top 10 post in "books" forum, show me the book names from posts that recommand a single book.
\end{tcolorbox}

\paragraph{Settings.}

We include five subtasks:  E-commence, Reddit, Gitlab, OpenStreetMap (Map), and online store content management system (CMS), comprising a total of $372$ training queries and $50$ testing queries. These are selected from the origin WebArena dataset, which contains $812$ queries across three categories: Information Seeking, Site Navigation, and Content \& Config. To facilitate efficient parallel rollout, we exclude the Content \& Config tasks, which involve insert, update and delete operations that change the state of the websites. We set the maximum number of agent-environment interactions to $15$ turns. For the SFT baselines, we set the learning rate to $1 \times 10^{-4}$. We employ GRPO as the main RL algorithm with a learning rate of $5 \times 10^{-7}$ and a KL cofficient of $1 \times 10^{-3}$. For each query, we sample 4 distinct trajectories using a temperature of $1.0$.

\subsection{Deep Search Scenario}

\paragraph{Tools and APIs.}

The deep search senario features a search engine–based environment equipped with specialized tools and APIs supporting the interaction with search engines. These APIs enable agents to dynamically generate search queries during the reasoning process, retrieve relevant information from external sources, and incorporate the retrieved information into subsequent reasoning steps. This setting allows agents to engage in complex reasoning processes that involve iterative searching and information integration, thereby enhancing their capability to solve intricate problems where external knowledge is essential.

A query case of Deep Search is shown below:

\begin{tcolorbox}[title = {Deep Search Example},breakable]
You must always reason inside \textless{}think\textgreater{}...\textless{}/think\textgreater{} first; if you lack knowledge, issue a \textless{}search\textgreater{}...\textless{}/search\textgreater{} and then stop; do not generate \textless{}information\textgreater{} or \textless{}answer\textgreater{} yet; wait for external input between \textless{}information\textgreater{}...\textless{}/information\textgreater{} before continuing; resume only when new \textless{}information\textgreater{} is given; do not skip steps or anticipate answers early.

\tcblower

Question: Who got the first Nobel Prize in Physics?
\end{tcolorbox}

\paragraph{Settings.}
We include queries from 7 datasets following the setup of Search-R1 \citep{Search-R1}: NQ \citep{NQ}, TriviaQA \citep{TriviaQA}, PopQA \citep{PopQA}, HotpotQA \citep{HotpotQA}, 2wiki \citep{2WikiMultiHopQA}, Musique \citep{Musique}, and Bamboogle \citep{Bamboogle}. To ensure fair comparison and balanced evaluation, we randomly sample 400 examples from the development sets of NQ, TriviaQA, PopQA, HotpotQA, 2wiki, Musique, and Bamboogle. The maximum number of agent-environment interactions is set to $4$ turns. For the SFT baselines, the learning rate is set to $1 \times 10^{-4}$. 
We employ GPRO as the main algorithm for reinforcement learning setups with a learning rate of $1 \times 10^{-6}$, a KL cofficient of $1 \times 10^{-3}$, and a sampling temperature of $1.0$. We sample $8$ distinct trajectories for a single query.

\subsection{Digital Games Scenario}
\paragraph{Environments, Tools and APIs.}
As for digital games, we introduce TextCraft\citep{ADaPT}, a text-based game environment mirroring Minecraft. The APIs in TextCraft include crafting, inventory management, and dynamic narrative generation. These APIs allow agents to execute predefined crafting recipes, manipulate inventory contents, navigate virtual spaces, dynamically generate quests and sub-tasks based on natural language objectives, and recursively decompose complex tasks into achievable sub-goals. 

A query case of TextCraft can be seen below:

\begin{tcolorbox}[title = {TextCraft Example},breakable]
You are given few useful crafting recipes to craft items in Minecraft. Crafting commands are of the format "craft [target object] using [input ingredients]".

Every round I will give you an observation, you have to respond an action based on the state and instruction. You can "get" an object (ingredients) from the inventory or the environment, look-up the game inventory by "inventory", or "craft" (target) using any of the crafting commands. You can use ONLY these crafting commands provided, do not use your own crafting commands. However, if the crafting command uses a generic ingredient like "planks", you can use special types of the same ingredient e.g. "dark oak planks" in the command instead.

\tcblower

Goal: Craft flint and steel.
\end{tcolorbox}

\paragraph{Settings.}
In TextCraft, task difficulty is measured by the maximum depth of the corresponding crafting tree. In practice, the benchmark contains tasks with crafting trees of depths 1, 2, 3, and 4. Accordingly, we divide the entire task set into four subsets based on these depths. We set the maximum number of interactions to $20$ turns. In the SFT baselines, we set the learning rate to $1 \times 10^{-4}$. We employ GRPO as the main RL algorithm with a learning rate of $1 \times 10^{-6}$, a KL cofficient of $1 \times 10^{-3}$, and a sampling temperature of $1.0$. We sample $8$ distinct trajectories for a single query.

\subsection{Embodied Scenario}

\paragraph{Tools and APIs.}
We introduce the BabyAI environment as a representative setting for embodied tasks.  It provides APIs that allow agents to navigate a controllable grid world using natural language instructions. Through these APIs, agents can perform actions such as moving objects, unlocking doors, and interacting with the environment in response to textual commands.

A query case of BabyAI can be seen below:

\begin{tcolorbox}[title = {BabyAI Example},breakable]
You are an exploration master that wants to finish every goal you are given. Every round I will give you an observation, and you have to respond an action and your thought based on the observation to finish the given task. You are placed in a room and you need to accomplish the given goal with actions.

You can use the following actions: 

- turn right - turn left - move forward - go to \textit{obj} \textit{id} - pick up \textit{obj} \textit{id}

- go through \textit{door} \textit{id}: \textit{door} must be an open door.

- toggle and go through \textit{door} \textit{id}: \textit{door} can be a closed door or a locked door. If you want to open a locked door, you need to carry a key that is of the same color as the locked door.

- toggle: there is a closed or locked door right in front of you and you can toggle it.

\tcblower

Your goal: Go to the red ball.
\end{tcolorbox}

\paragraph{Settings.}
Following the original implementation, we divide the tasks into six subsets based on the final goal. We set the maximum number of interactions  to $20$ turns. In SFT baselines, we set the learning rate to $1 \times 10^{-4}$. We employ GRPO as the main RL algorithm with a learning rate of $1 \times 10^{-6}$, a KL cofficient of $1 \times 10^{-3}$, and a sampling temperature of $1.0$. We sample $8$ distinct trajectories for a single query.

\subsection{Scientific Scenario}
\paragraph{Tools and APIs.}
SciWorld\citep{ScienceWorld} is an agent environment for scientific tasks. It provides APIs that are designed to support scientific exploration through text-driven reasoning cycles. These APIs empower agents to conduct experiments by interacting with various scientific apparatus and performing actions like measuring temperature, connecting electrical circuits, and mixing chemicals.

A query case of SciWorld can be seen below:

\begin{tcolorbox}[title = {SciWorld Example},breakable]
You are an agent for science world. Every round I will give you an observation, you have to respond an action based on the observation to finish the given task.

\tcblower

Your task is to boil water. For compounds without a boiling point, combusting the substance is also acceptable. First, focus on the substance. Then, take actions that will cause it to change its state of matter.
\end{tcolorbox}

\paragraph{Settings.}
We select 8 subsets of tasks from the original SciWorld environment. We set the maximum number of interactions between the agent and the environment to $20$ turns. In SFT baselines, we set the learning rate to $1 \times 10^{-4}$. We employ GRPO as the main RL algorithm with a learning rate of $1 \times 10^{-6}$, a KL cofficient of $1 \times 10^{-3}$, ans a sampling temperature of $1.0$. We sample $8$ distinct trajectories for a single query.

\section{Trajectory examples and visualizations of our RL agent}
\label{appendix:trajectory}

This appendix provides additional trajectory visualizations across multiple environments. The figures illustrate the behaviors of both baseline and RL-trained agents, highlighting the RL model’s superior performance in exploration, task execution, and interaction patterns, while also revealing common failure modes that remain.

\begin{figure}[t]
    \centering
    \includegraphics[width=\linewidth]{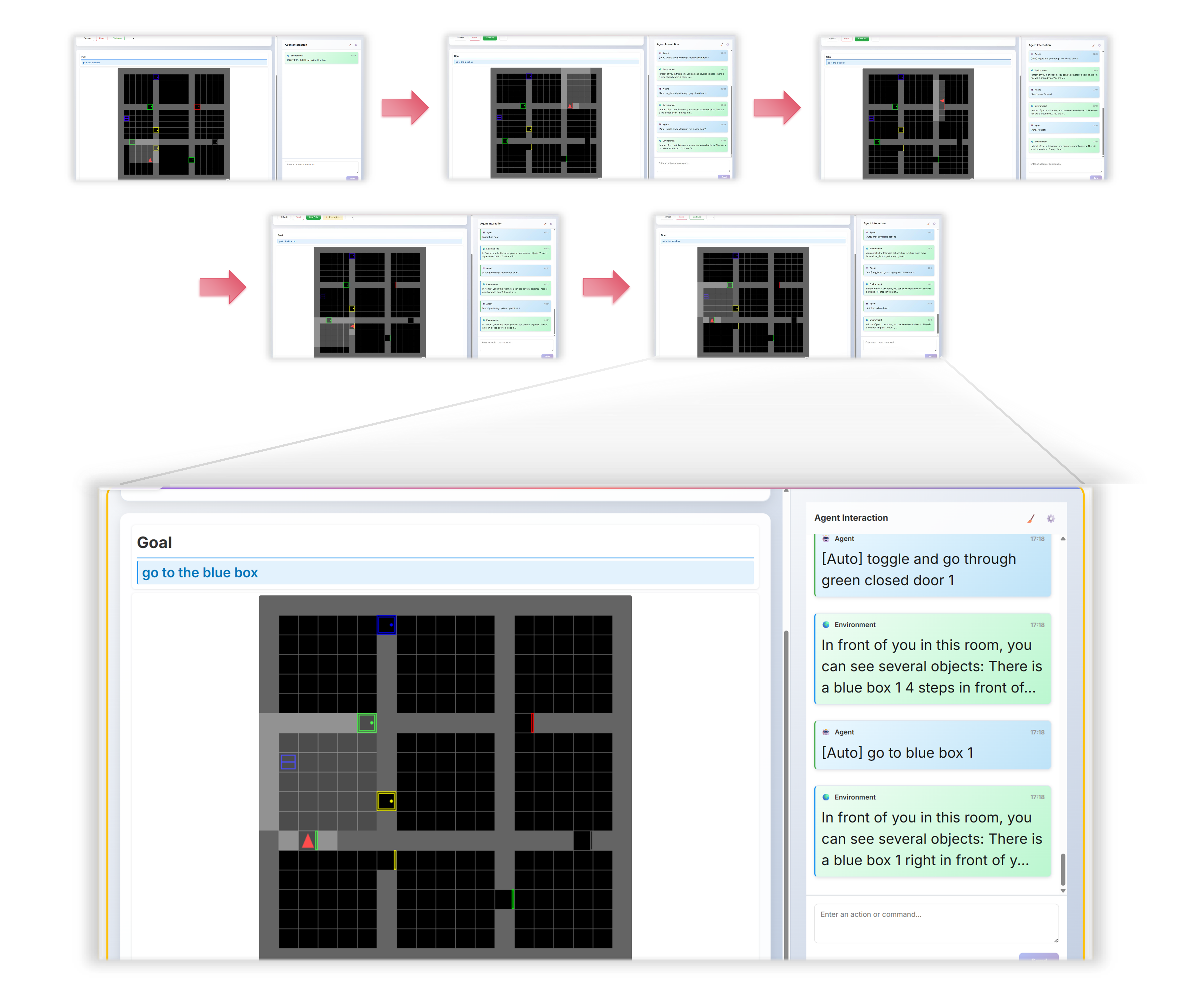}
    \caption{
    Trajectory visualization of our RL agent on BabyAI task.
    }
    \label{fig:case_babyai_positive_visualization}
\end{figure}

\end{document}